\documentclass[5p,twocolumn]{elsarticle}

\usepackage{seqsplit}
\usepackage{amsmath}
\usepackage{enumitem}
\usepackage{graphicx}
\usepackage{algorithmic,algorithm}
\usepackage{epstopdf}
\usepackage[justification=centering]{caption}
\usepackage{color}
\usepackage{picins}
\setlist{nolistsep}

\makeatletter
\def\ps@pprintTitle{%
  \let\@oddhead\@empty
  \let\@evenhead\@empty
  \let\@oddfoot\@empty
  \let\@evenfoot\@oddfoot
}
\makeatother

\usepackage{amssymb}
\usepackage{latexsym}

\begin{document}

\begin{frontmatter}

\title{Self-taught learning of a deep invariant representation for visual tracking via temporal slowness principle}

\author{Jason Kuen}
\ead{jason7fd@gmail.com}

\author[fistmmu]{Kian Ming Lim\corref{mycorrespondingauthor}}
\cortext[mycorrespondingauthor]{Corresponding author. Tel.: +606 2523066; fax.: +606 2318840.}
\ead{kmlim@mmu.edu.my}

\author[fistmmu]{Chin Poo Lee}
\ead{cplee@mmu.edu.my}

\address[fistmmu]{Faculty of Information Science and Technology, Multimedia University, Malaysia}

\begin{abstract}
Visual representation is crucial for a visual tracking method's performances. Conventionally, visual representations adopted in visual tracking rely on hand-crafted computer vision descriptors. These descriptors were developed generically without considering tracking-specific information. In this paper, we propose to learn complex-valued invariant representations from tracked sequential image patches, via strong temporal slowness constraint and stacked convolutional autoencoders. The deep slow local representations are learned offline on unlabeled data and transferred to the observational model of our proposed tracker. The proposed observational model retains old training samples to alleviate drift, and collect negative samples which are coherent with target's motion pattern for better discriminative tracking. With the learned representation and online training samples, a logistic regression classifier is adopted to distinguish target from background, and retrained online to adapt to appearance changes. Subsequently, the observational model is integrated into a particle filter framework to peform visual tracking. Experimental results on various challenging benchmark sequences demonstrate that the proposed tracker performs favourably against several state-of-the-art trackers.

\end{abstract}


\end{frontmatter}


\section{Introduction}
\label{sec1}
Visual tracking is one of the most important research topics in computer vision because it is in the core of many real-world applications. Applications of such include human-computer interactions, video surveillance, and robotics. Due to the need for generality, recent years have seen the rise of online model-free visual tracking methods which attempt to learn the appearance of the target object over time, without prior knowledge about the object. Despite much research efforts have been made, visual tracking is still regarded as a challenging task due to various appearance changes of the target object and background distractions. Illumination variations, occlusion, fast motion, and background clutters are some challenges in visual tracking.

A typical visual tracking method is dependent on its two major components \cite{li2013asurvey}, namely dynamic model (motion estimation) and observational model. A dynamic model is used to model the states and state transition of the target object, whereas an observational model describes the target object and observations based on certain visual representations. To deal with the abovementioned visual tracking challenges, most recent tracking methods tend to put focus on adopting or developing more effective representations. However, variants of image representations (e.g., Histogram of Oriented Gradients (HOG), Scale Invariant Feature Transform (SIFT), Local Binary Patterns (LBP)) developed in the computer vision domain are not universally effective on wide-range of vision tasks, and they lack of customizability. One recent and highly effective approach to have better task-specific representations, is to learn representations from raw data itself. Representation learning techniques seek to bypass the conventional way of labor-intensive feature engineering, by disentangling the underlying explanatory factors for the observed input. Thus, representation learning will be the main focus of our approach.

Objects in a video are likely to be subject to small transformations across frames but the content remains largely unchanged. Our work presented in this paper aims to exploit temporal slowness principle to learn an image representation which change slowly over time, thus making it robust against these local transformations. Making use of a big amount of unlabeled tracked sequential data, generic local features invariant to transformations commonly found in tracking tasks can be learned offline. To that end, a complex-cell-like autoencoder model with temporal slowness constraint is proposed for learning separate representations of invariances and their transformations in the image sequences. To learn more complex invariances, a deep learning model  is formed by training a second autoencoder with the convolved activations of the first autoencoders on larger image patches.

\begin{figure*}[!htb]
\centering
\includegraphics[scale=.5]{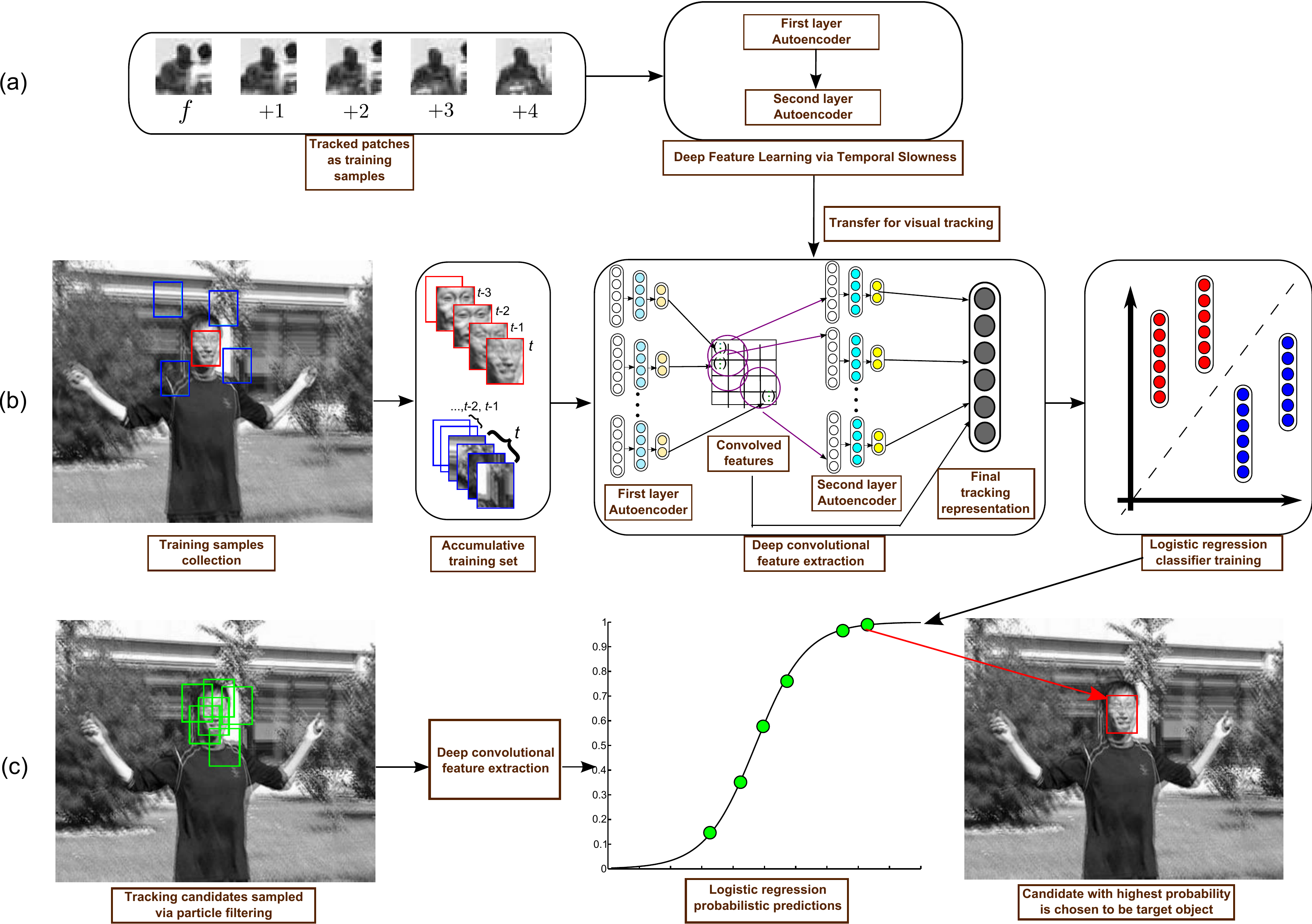}
\caption{Overview of proposed tracker in terms of three major stages: (a) offline learning of deep slow representations, (b) observational model update, and (c) tracking.}
\label{fig:flow}
\end{figure*}

The overview of our proposed method with its three major components is illustrated in Fig.\ \ref{fig:flow}. Firstly, in Fig.\ \ref{fig:flow}(a), tracked image patches are used to train the deep stacked autoencoders via temporal slowness constraint (refer to Fig.\ \ref{fig:autoencoder} for visual details). The trained stacked autoencoders are then transferred to an adaptive observational model for visual tracking (Fig.\ \ref{fig:flow}(b)(c)). Based on certain conditions during tracking, the observational model is updated online to account for appearance changes. Fig.\ \ref{fig:flow}(b) describes the steps for observational model update, whereby logistic regression classifier is trained on an accumulative training set, with the features obtained from the transferred stacked autoencoders. In Fig.\ \ref{fig:flow}(c), tracking is performed by sampling tracking candidates via particle filtering. With the learned representation and trained logistic regression, the candidate with the highest predicted probability is chosen to be the target object.

The main contributions of this paper are:
\begin{enumerate}
  \item We present an autoencoder algorithm to learn generic invariant features offline for visual tracking. To train the model, we perform tracking on unlabeled sequential data and obtain tracked image patches as training data. Transformation-invariant features are learned by enforcing strong temporal slowness between tracked image patches. With subspace pooling, we construct a complex-valued representation which separates invariances from their transformations.  We further add another autoencoder layer to construct a stacked convolutional autoencoders model for learning higher-level invariances. The stacked autoencoders are then transferred for use in visual tracking, based on self-taught learning paradigm \cite{raina2007self}.
  \item With the learned representations, we propose an adaptive observational model for tracking. Both first and second layer of the stacked autoencoders are transferred to form a final tracking representation. For better discriminative tracking, the proposed observational model is equipped with a novel negative sampling method which collects more relevant negative training samples. Besides, to alleviate visual drift, we propose a simple technique for the observational model to retain early and recent training samples.
  \item We integrate the proposed adaptive observational model into a particle filter framework and evaluate our proposed tracker on a number of challenging benchmark sequences, comparing with several state-of-the-art trackers. Results demonstrate that the proposed tracker performs favourably against the competing trackers.
\end{enumerate}

\section{Related work}
Observational model, also known as appearance model is undoubtedly the most crucial component in visual tracking. In this section, literature review is done for existing trackers in terms of the two common categories of observational model, namely generative and discriminative approaches. Subsequently, several existing representation learning-based trackers are reviewed.

Generative approaches represent target object with low reconstruction error and identifies the best matched candidate among many observations. To adapt to appearance changes of target object, most of the recent generative tracking methods learn the appearance of the object online. Subspace learning methods learn expressive representations in low dimensional space. To develop a subspace learning-based tracker, \citet{ross2008incremental} used Principal Component Analysis (PCA) to construct and incrementally update a subspace model of the target object. \citet{liwicki2012efficient} formulated an incremental kernel PCA in Krein space to learn a nonlinear subspace representation for tracking. To account for partial occlusion during tracking, \citet{kim2012correlation} proposed a Canonical Correlation Analysis (CCA)-based tracker which considers the correlations among sub-patches of tracking observations. Mixture models can also be used for tracking, \citet{jepson2003robust} learned a mixture model to model appearance changes of target object via an online Expectation-Maximisation (EM) algorithm. \citet{wang2007adaptive} develop an adaptive observational model in the joint spatial-color space using Gaussian mixture model.

Although generative trackers can work well in certain circumstances, they are inferior to discriminative trackers in dealing with complicated environments.  Unlike generative trackers, discriminative trackers take background information into account and distinguish between target object and background. \citet{collins2005online} selected color features online which best discriminates target object from current background. To deal with appearance changes, \citet{grabner2006real} proposed an online boosting classifier that adaptively selects discriminative features for tracking.  \citet{klein2011boosting} introduced a novel scale-invariant gradient feature and used boosting to track target object efficiently. To alleviate visual drift, \citet{zhang2013owmil} proposed a tracking method based on online multiple instance boosting that handles ambiguously labeled samples and weights each positive sample differently for update.  Besides online feature selection and ensemble methods, Support Vector Machine (SVM)-based tracking methods have received much attention lately. \citet{tang2007cotracking} trained multiple SVMs, each on an independent feature and locates the target object by combining confidence scores from all SVM classifiers.  To reduce the error of selecting inaccurate samples in updating observational model online, \citet{hare2011struck} presented a structured output SVM to directly predict the trajectory of the target object between frames. Discriminative tracking methods are advantageous because they generally allow the incorporation of various kinds of feature representations. However, if these representations are not well formulated, they can be devastating for the tracking performances.

Representation learning \citep{bengio2013repres} is an emerging field aims to learn good representations from raw input or low-level representations. Many representation learning techniques (e.g., \citet{netzer2011reading}; \citet{yu2011learning}) have proven to be superior to conventional hand-engineered representations. Two popular techniques in representation learning are dictionary learning and deep learning. Dictionary learning aims to learn a dictionary or codebook in which only a few atoms can be linearly combined to well approximate a given signal. Considering the absence of object prior information in existing model-free tracking methods, \citet{wang2012transferring} performed sparse coding on SIFT features extracted from labeled object recognition datasets. \citet{liu2013online} learned a sparse coded dictionary online from raw unlabeled patches and discriminates target object from background regions using SVM. Deep learning is relatively new in visual tracking research. In deep learning, deep layered architectures are employed to learn complex high-level representations in terms of less simpler low-level representations.  Using \textit{k}-means clustering, \citet{jonghoon2013tracking} trained a convolutional neural network offline in an unsupervised manner for tracking. Their method however only maintains a static observational model over time. To account for object appearance changes, \citet{wang2013learning} pre-trained a stacked denoising autoencoders and fine-tuned the deep neural network online during tracking.

\section{Learning deep invariant representations using temporal slowness }

In this paper, we aim to exploit temporal slowness to learn generic invariant representations in an unsupervised way and transfer them for visual tracking. Temporal slowness is one of the major priors for representation learning  \citep{bengio2013repres} and it has been successfully used in object recognition tasks (e.g., \citet{mobahi2009deep}; \citet{zou2012deep}).  The main motivation of learning such representation is that it is difficult to develop or hand-craft an exact feature extraction algorithm that is robust against object transformations found in videos. To the best of knowledge, this paper is the first attempt to employ temporal slowness principle for visual tracking. \citet{zou2012deep} showed that temporal slowness constraint can be simply added to conventional autoencoder cost function to learn invariant features, we also choose autoencoder to be our representation learning model. Although our proposed tracker is discriminative in nature, the representations are learned via generative constraints.

This section is organized as follows: Firstly, the conventional autoencoder is introduced and modifications are progressively added to the autoencoder to achieve the temporally slow autoencoder algorithm. Secondly, the details on how a stacked convolution autoencoder can be constructed are described. Thirdly, the complex-valued representation formed by the subspace pooled autoencoder is described and the visualized optimal stimuli for the learned features are shown. Lastly, the sequential dataset used and the preprocessing method to obtain tracked training image patches are explained.

\subsection{Autoencoder with temporal slowness constraint}
An autoencoder is an unsupervised artificial neural network which learns a mapping to reconstruct input data in its final layer. Given an input vector of $ N $ number of $ d_1 $-dimensional data samples $ [\pmb{x}^{(1)},...,\pmb{x}^{(N)}] \in \mathbb{R}^{d_1\times N} $, an autoencoder has a squared error cost function :
\begin{equation}\label{eq:1}
\sum_{i=1}^N
\| \pmb{x}^{(i)}-f_d(W_df_e(W_e\pmb{x}^{(i)}+\boldsymbol{\beta}_e)+\boldsymbol{\beta}_d) \|_2^2
\end{equation}

\noindent where $ W $ is the autoencoder weights, $ f $ is the activation function, $ \boldsymbol{\beta} $ is the network biases, and subscripts $ e $ and $ d $ indicate the associations of the components with the encoder and decoder respectively. Generally, the latent representation $ f_e(W_e\pmb{x}^{(i)}+\boldsymbol{\beta}_e) $ learned in the hidden layer of an autoencoder is regarded to be meaningful for classification purposes. The conventional autoencoder with mere reconstruction cost can hardly learn any useful latent representation of the data. To allow autoencoder to discover better representations, many autoencoder regularization schemes such as denoising autoencoders \citep{vincent2008extracting} and contractive autoencoders \citep{rifai2011contractive} have been proposed. However, these autoencoder variants use only static and temporally uncorrelated object images to learn features for object recognition tasks. Although such approach can be borrowed directly for visual tracking purpose (e.g., \citet{wang2013learning}), we contend that image representation for visual tracking should be learned in a way more specific to the task.

\begin{figure}[!htb]
\centering
\includegraphics[scale=.5]{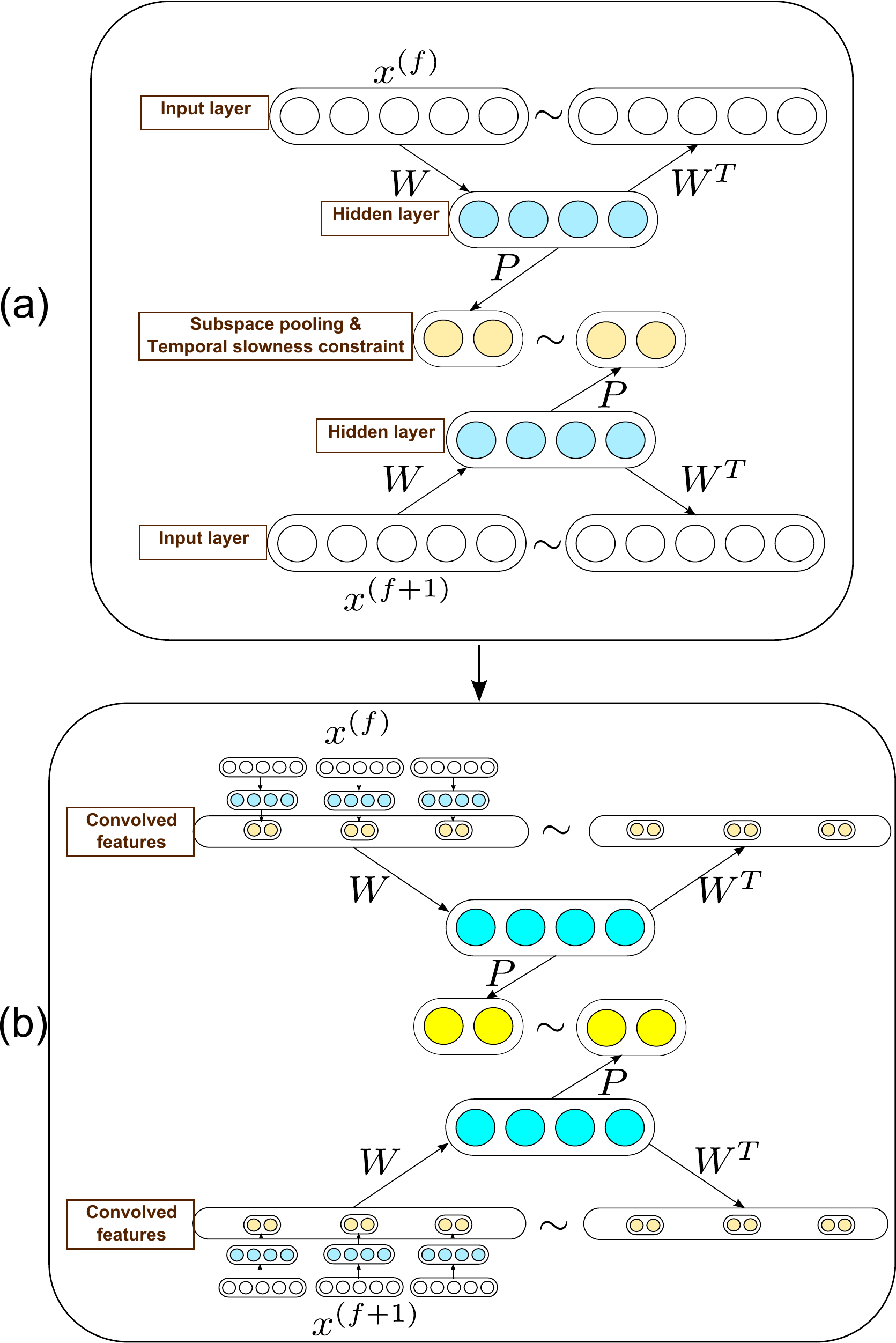}
\caption{Architecture of stacked convolutional autoencoder with subspace pooling (blue) and temporal slowness (yellow), (a) first layer and (b) second layer.}
\label{fig:autoencoder}
\end{figure}

To this end, we propose to adapt autoencoder to learn invariant features from tracked image sequences for visual tracking. Our autoencoder model draws inspiration from Independent Subspace Analysis (ISA) \citep{le2011learning} which was proposed for learning motion invariance. Unlike conventional sparse autoencoders which enforce sparsity in the hidden layer, the proposed autoencoder performs subspace pooling on the hidden layer activations and enforces sparsity in the pooling layer, in a way identical to ISA. When learning features, invariance is achieved by enforcing a strong temporal slowness constraint to minimize the distance between the subspace pooling representations of any two temporally correlated tracked image patches. To preserve the architectural properties of ISA that it has no biases and nonlinear activation function, biases are omitted and linear activation function is chosen in the reconstruction cost in Eq.\ (\ref{eq:1}). Furthermore, the encoder and decoder weights are tied $ (W_d=W_e^T) $ to reduce the number of free parameters to train. In the proposed autoencoder, the reconstruction cost replaces hard orthonormality constraint in ISA to prevent feature degeneracy \citep{le2011ica} and the sparsity cost helps to discover interesting features \citep{glorot2011deep} (e.g., edges, corners).

To train the autoencoder, we use a dataset consists of $ N $ number of tracked patches, formed by $ N^T $ number of track sessions and $ N^F $  number of frames per track (refer to Section 3.4). The modified autoencoder with $ p $ number of hidden units is then trained by :
\begin{multline}\label{eq:2}
\sum_{i=1}^N
\| \pmb{x}^{(i)}-W^TW\pmb{x}^{(i)} \|_2^2
+
\alpha
\sum_{t=1}^{N_T}
\sum_{f=1}^{N_F-1}
\| \pmb{h}^{(t,f)}-\pmb{h}^{(t,f+1)} \|_1 \\
+
\gamma
\sum_{i=1}^N
\| \pmb{h}^{(i)} \|_1
\end{multline}

\noindent
where $ W \in \mathbb{R}^{p\times d_1} $ is the autoencoder weights, $ \alpha \in \mathbb{R} $  is the weight of the temporal slowness cost, and $ \pmb{h}^{(t,f)} \in \mathbb{R}^{(\frac{p}{2})\times 1} $  is the subspace pooling representation of image patch at $ f $-th frame of $ t $-th track.  $ L^1 $-norm minimization is a common way to achieve sparsity in representation learning. In the second cost term (temporal slowness constraint), as in \citep{mobahi2009deep} and \citep{zou2012deep},  we minimize the temporal representation differences in $ L^1 $-norm to allow invariance to be sparsely represented, that is a kind of motion invariance is represented by only a small number of features and thus they become specialized for different invariances. $ L^1 $-norm regularization is too applied to the third cost function term to enforce sparsity in the subspace pooling layer $ \pmb{h} $, where $ \gamma \in \mathbb{R} $ parameterizes the weight of the sparsity regularization.

The $ i $-th subspace pooling unit $ \pmb{h}^{(i)} $ is obtained by performing $ L^2 $-norm subspace pooling on its hidden layer counterpart :
\begin{equation}\label{eq:3}
\pmb{h}^{(i)} = \sqrt{P(W\pmb{x}^{(i)})^{\cdot 2}}
\end{equation}

\noindent
where $ ()^{\cdot 2} $ indicates element-wise square operation and $ P \in \mathbb{R}^{(\frac{p}{2})\times p} $ is a subspace pooling matrix which sums up every two adjacent features in a non-overlapping way. The encoder diagram with subspace pooling is shown in Fig.\ \ref{fig:autoencoder}(b). Subspace pooling has been successfully used in temporal slowness feature learning techniques (e.g., \citet{bengio2009slow}; \citet{zou2012deep}) to group similar features in each of the pooled units, therefore achieving invariance. It pairs every 2 adjacent hidden units to form a complex-valued representation and each pair can be decomposed into amplitude (degree of presence of the features) and phase (transformations of the features over time) variables \citep{olshausen2007bilinear}. The pooling units resemble complex-cells in the visual cortex, in which the amplitudes are insensitive to phase changes. The proposed autoencoder with temporal slowness constraint on any two paired tracked patches is illustrated in Fig.\ \ref{fig:autoencoder}(a).

Unlike ISA which relies on inconvenient constrained optimization methods for training, the cost function in Eq.\ (\ref{eq:2}) can be optimized efficiently using any of the unconstrained optimization methods \citep{ngiam2011optimization}.

\subsection{Stacked convolutional autoencoders}
Autoencoders have been commonly stacked to form deep layered architectures to learn higher-level representations from its low-level counterparts. In this paper, we stack and train a second autoencoder (known as the second layer) on the convolved features of the first autoencoder (known as the first layer) in a greedy layer-wise training fashion \citep{larochelle2009exploring}. The convolutional learning architecture allows the reuse of first layer features to learn higher-level features from bigger image patches.

Before training the second layer, the first layer subspace pooling features are densely extracted from larger $ d_2 $-dimensional tracked image patches (where $ d_2 > d_1 $) with a predefined spatial stride or step-size $ k_1 $, where $ k_1 \geq 1 $. Experimentally, we choose $ k_1 $ in such a way that the overlaps between the dense patches is small. A small $ k_1 $ is computationally demanding, while a big $ k_1 $ with no overlap impedes the feature performances severely. Unlike in object recognition tasks whereby a small $ k_1 $ is always recommended to achieve good performances \citep{coates2011analysis}, there is a great need to balance the feature and run-time performances in visual tracking.

\begin{figure}[!htb]
\centering
\includegraphics[scale=1.15]{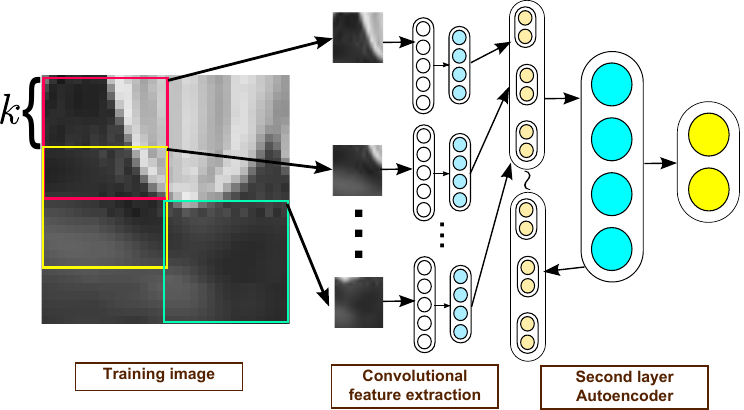}
\caption{Convolutional feature extraction for training the second layer.}
\label{fig:convolution}
\end{figure}

Subsequently, the convolved first layer representation from the densely extracted patches are flattened into a feature vector per large tracked image, and they are used as training dataset to train the second layer. The convolutional learning architecture of this method is illustrated in Fig.\ \ref{fig:autoencoder}(b) and \ref{fig:convolution}. Due to a bigger $ k_1 $ chosen, we get a reasonable number of feature dimensions in the convolved representations, and thus a dimensionality reduction technique is not required unlike the strategy of \citet{zou2012deep}.  The second layer shares the same autoencoder cost function as the first layer but they differ in parameter settings for the cost term weights.

\subsection{Feature extraction and visualization}
As pointed out in Section 3.1, there are two kinds of information can be obtained from the subspace pooled units of the proposed autoencoder. Let $ A \in \mathbb{R} $ and $ B \in \mathbb{R} $ denote any paired adjacent activations in the hidden layer of the autoencoder, a complex representation $ A+ iB \in \mathbb{C} $ is formed. The amplitude or magnitude of the complex representation $ A+ iB $ indicates the degree of presence of the feature while being invariant to its phase changes and transformations. It is computed as the Euclidean norm of the complex number :

\begin{equation}\label{eq:4}
\sqrt{A^2+B^2}
\end{equation}

\noindent
which is exactly what Eq.\ (\ref{eq:3}) does. The amplitudes are a good invariant representation for supervised classifications. The second information obtained from the subspace pooled units is the phase. Phase of the complex representation $ A+ iB $ is defined as :

\begin{equation}\label{eq:5}
\tan^{-1}\left(\frac{B}{A}\right) .
\end{equation}

\noindent
With temporal slowness constraint, the angular-valued phase models the feature transformations of each pooled unit in smooth transitions, which means that the features change slowly with gradual change in phase. For visual tracking, both amplitude and phase features are considered due to their invariant and discriminative properties respectively.

\begin{figure}[!htb]
\centering
\includegraphics[scale=.425]{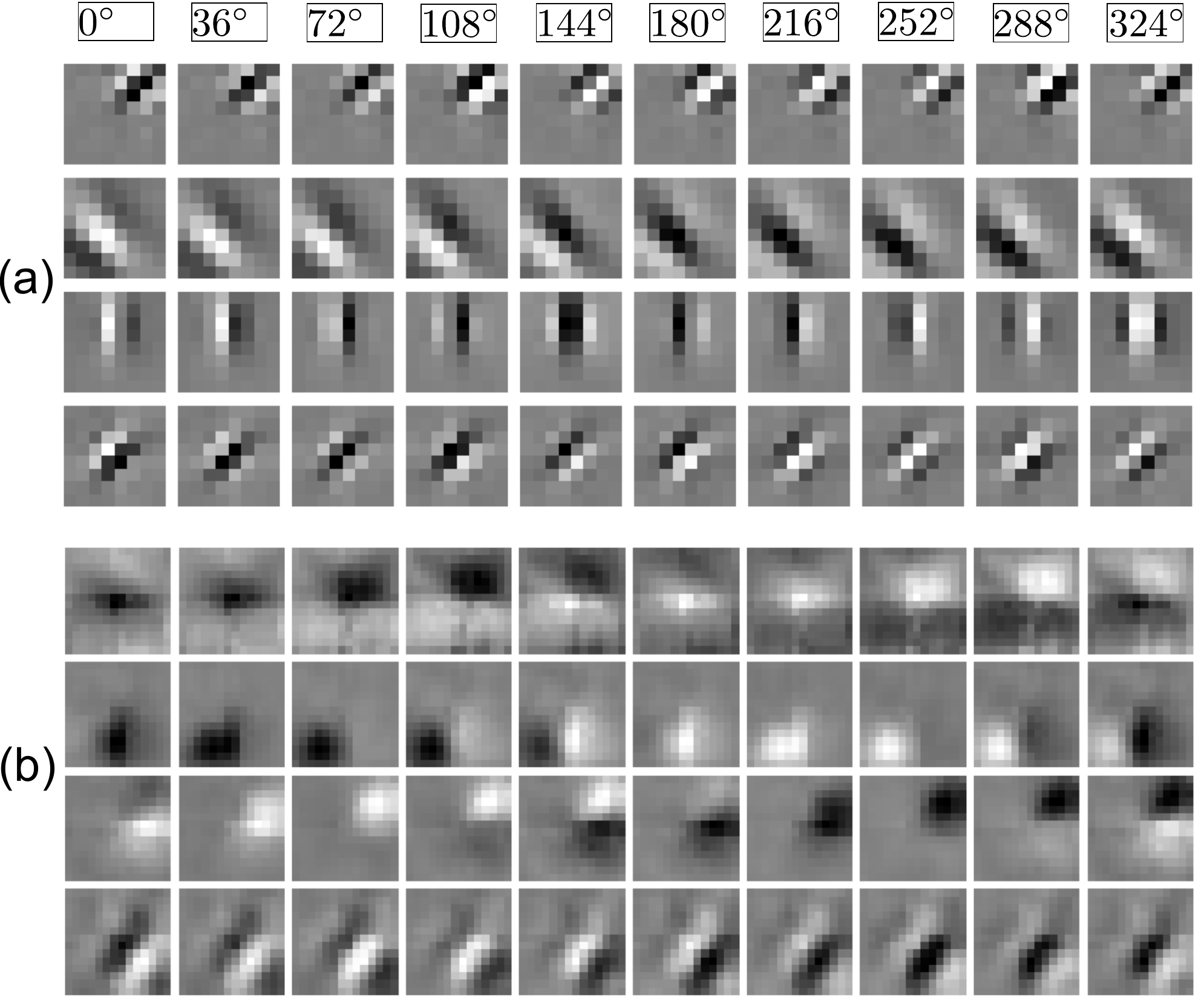}
\caption{Visualization of (a) first-layer and (b) second-layer optimal stimuli and invariances.}
\label{fig:vis_1}
\end{figure}

An interesting property of learning with subspace pooling and temporal slowness is that it allows the phase of the pooled units to be shifted (\citet{olshausen2007bilinear}; \citet{zou2012deep}), to visualize what transformations the features are invariant to. We follow the phase-shifting procedure of \citet{zou2011unsupervised} and use linear combination of filters \citep{erhan2009visualizing} to visualize the optimal stimuli for the features and invariances learned in the stacked autoencoders. Fig.\ \ref{fig:vis_1} shows some representative optimal stimuli for the two layers of our stacked autoencoders, with a $ 36\,^{\circ} $ increment in the phase shift. It is shown that the first layer learns Gabor-like edge detectors which are invariant against small location translations, and the second layer learns features which are invariant against complicated transformations such as out-of-plane rotation.

\begin{figure}[!htb]
\centering
\includegraphics[scale=.425]{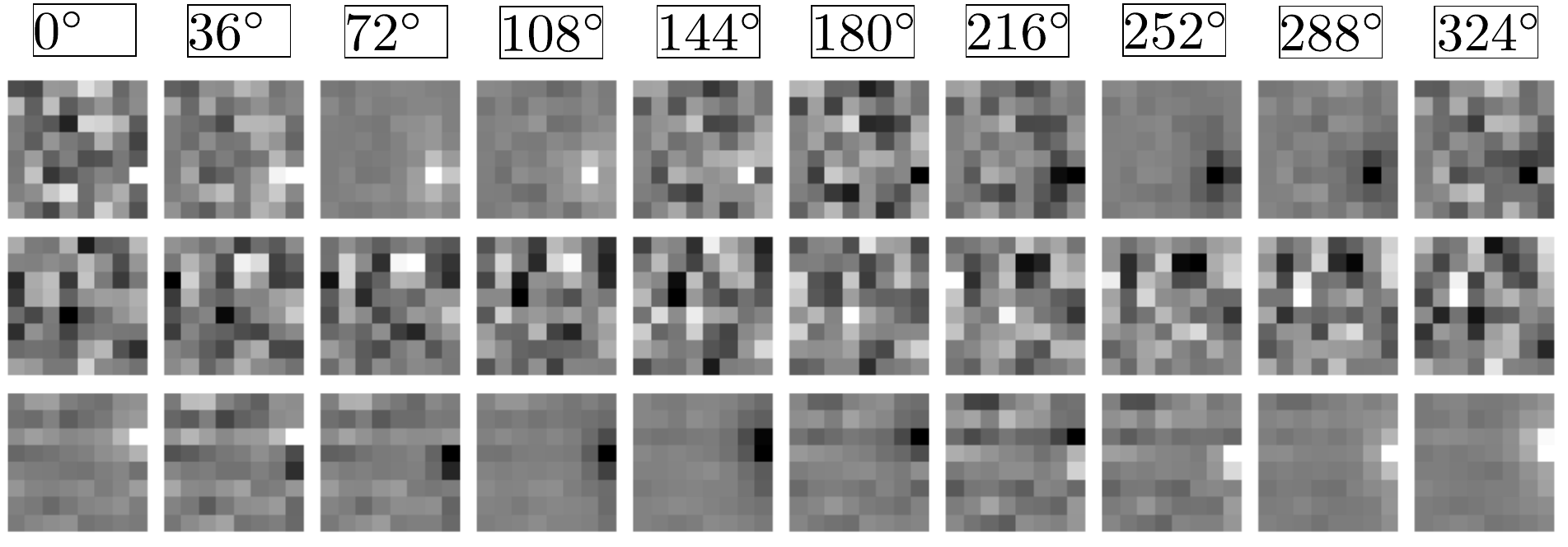}
\caption{Visualization of first-layer optimal stimuli and invariances, learned using temporally uncorrelated tracked patches.}
\label{fig:vis_bad}
\end{figure}

\subsection{Sequential training data collection}
We employ self-taught learning principle to learn a representation for visual tracking. In self-taught learning \citep{raina2007self}, features are learned from unlabeled data and transferred for use in supervised learning tasks, whereby the generating distribution of the unlabeled data is different from the labeled data. In the context of this paper, features are learned from image sequences unrelated to tracking benchmark sequences used to gauge the proposed tracker's performances. This setting is analogous to the tracking algorithm \citep{wang2012transferring} which exploits patch-level similarity and transfers visual prior from unlabeled dataset to tracking tasks.

Harnessing a diverse unlabeled dataset for self-taught feature learning is essential because the learned features are expected to generalize well to unseen examples. The deep learning-based tracking algorithm \citep{wang2013learning} which transfer features learned from unlabeled datasets, employ a very large dataset with great amount of visual diversity. In this paper, we use a large aggregation of tracking benchmark sequences compiled by \citet{wu2013online}\footnote{http://visual-tracking.net} and \citet{klein2010adaptive}\footnote{http://www.iai.uni-bonn.de/~kleind/tracking}, in which the objects and scenes are diverse in terms of their appearances. More importantly, they exhibit possibly all variants of tracking challenges such as occlusion, out-of-plane rotation, deformation, and scale variation. The experimental benchmark sequences used in Section \ref{experimental_setups} are excluded to uphold the self-taught learning principle.

By employing a tracker developed by \citet{jia2012visual}, tracked image patches are collected randomly from the sequences. The chosen tracker is not an arbitrary choice, it is the best performing tracker among the trackers we evaluated on a set of tracking benchmark sequences in Section \ref{experimental_setups}, excluding our own proposed tracker. To learn feature invariances via temporal slowness, it is important that the tracked image patches are accurately obtained. As a way to affirm this hypothesis, we randomly shuffled the first half of the our collected tracked dataset to disrupt the temporal ordering, and subsequently adopted it for learning first-layer features. This can be thought as a rough simulation of using a weak performing tracker to collect the tracked patches. The optimal stimuli for a few of the learned first-layer features are shown in Fig.\ \ref{fig:vis_bad}. It can be seen that the optimal stimuli are made up of noisy patterns and they do not resemble the sharp edge detectors as in Fig.\ \ref{fig:vis_1}.

\begin{figure}[!htb]
\centering
\includegraphics[scale=.425]{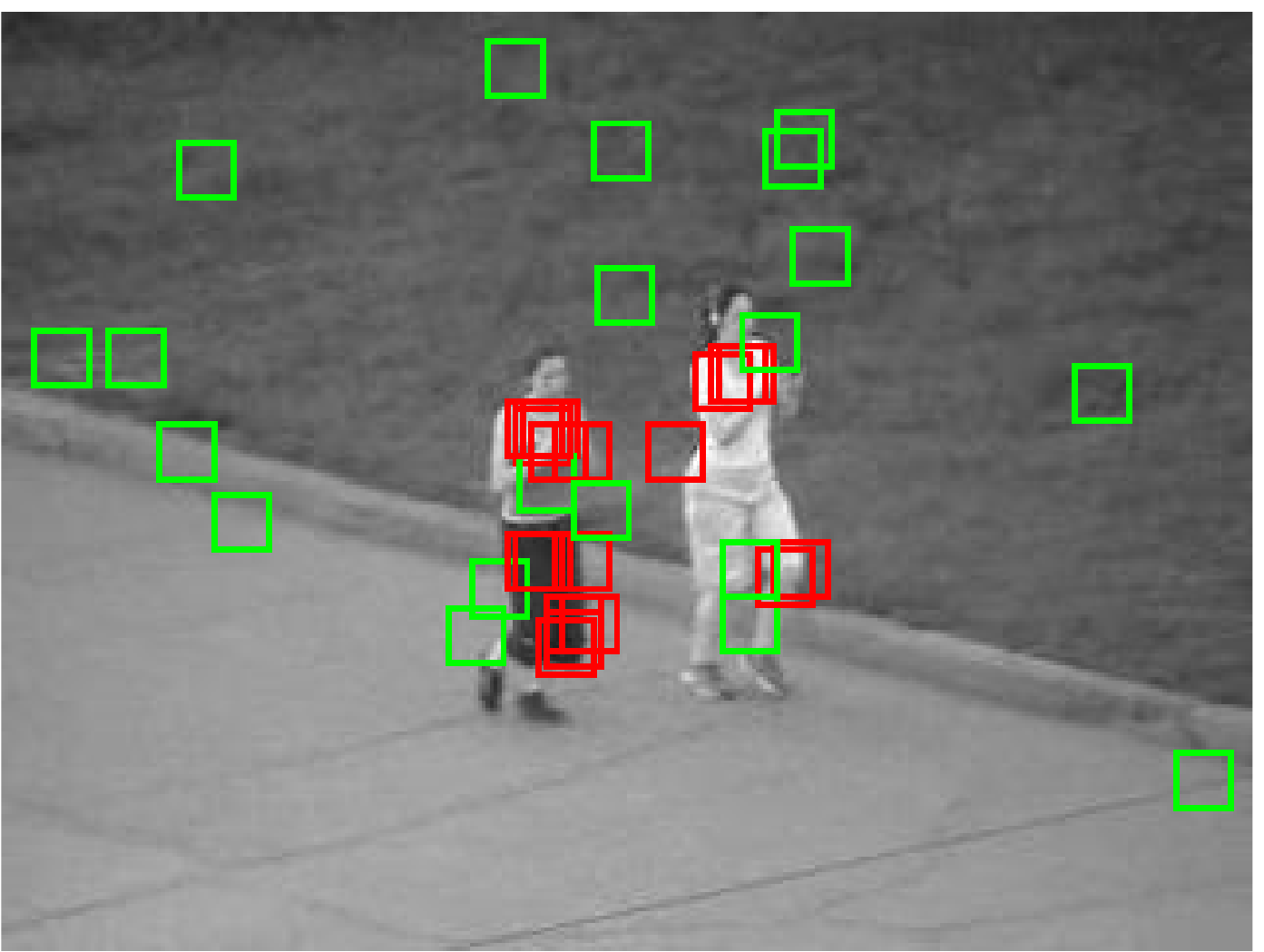}
\caption{Interest point detection comparison using \textit{accumulative difference pictures} (green) and \textit{STIP} (red).}
\label{fig:stip_running}
\end{figure}

Due to the presence of much uninteresting regions (e.g., flat appearance, appearance that remains constant over time) in the sequences, we find `interesting' regions before performing tracking. For interest point detection, two approaches are first considered. The first approach is about identifying motional pixels via binary-thresholded accumulative difference pictures \citep{jain1979analysis}, and a size filter is used to remove trivially small connected components among the identified pixels. Subsequently, initial tracking regions are chosen from random frame numbers and random spatial locations, with the constraint that the initial regions must have at least a small overlaps with the motional pixels. In contrast, the second approach involves a space-time Harris interest point detection algorithm \citep{laptev2005on} (referred to as STIP) that identifies regions which are `interesting' spatially and temporally. We then qualitatively compare the results of these two approaches on an arbitrary short video segment, as shown in Fig.\ \ref{fig:stip_running}. The video segment contains two running persons with a relatively unchanged background. Noticeably, STIP outperforms the first approach because the latter takes into account only temporal differences while ignoring much of regions which are rich in spatial information (e.g., corners, textures). Experimentally, we choose STIP over accumulative difference pictures because spatial and temporal interestingnesses help to learn good features and good invariances respectively.

\begin{figure}[!htb]
\centering
\includegraphics[scale=.425]{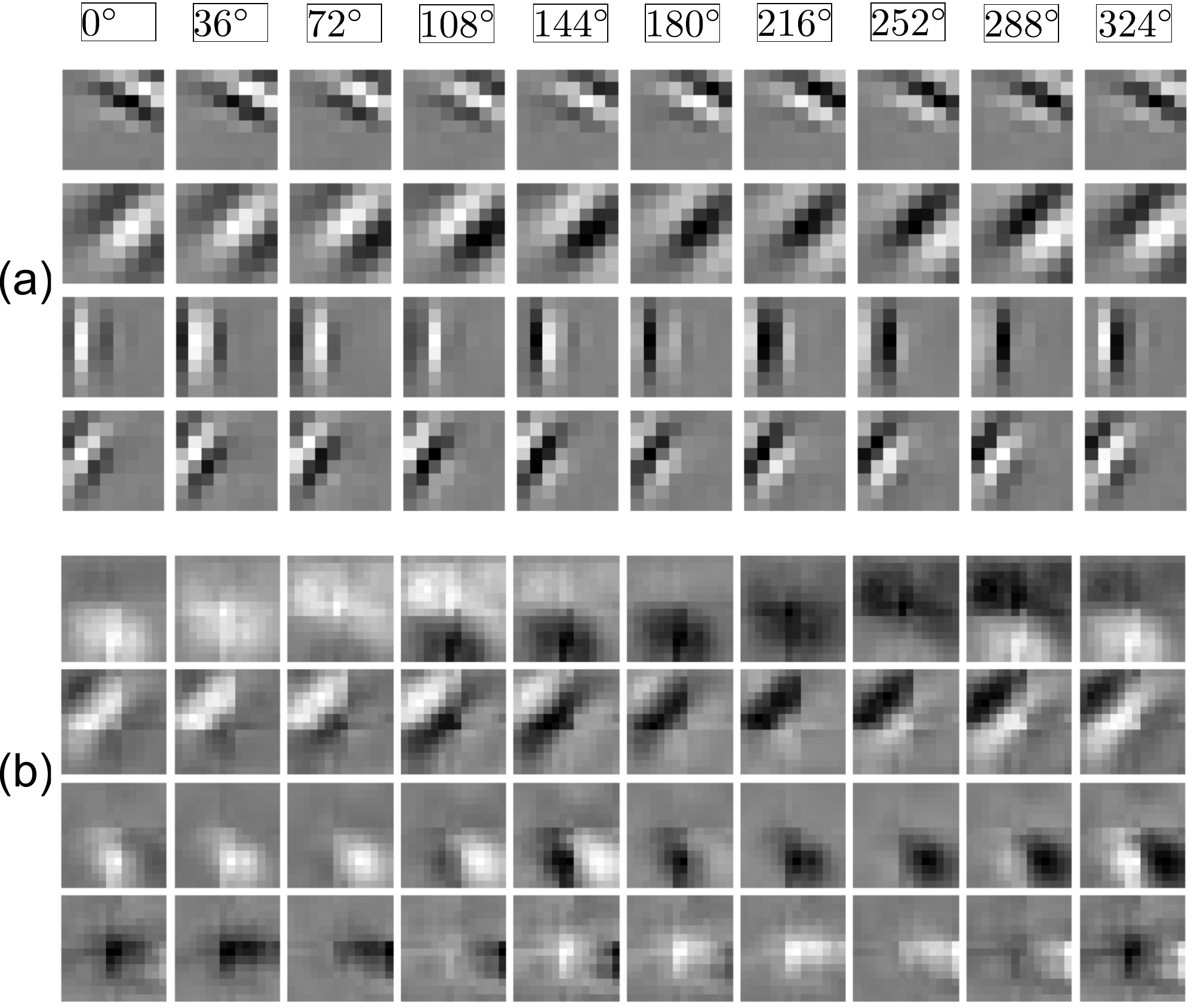}
\caption{Visualization of first-layer optimal stimuli and invariances, learned in \textit{20 frames/track} setting.}
\label{fig:vis_20}
\end{figure}

Compared to the work of \citet{zou2012deep}, we use a larger number of track sessions to encourage diversity in the training dataset, and use a smaller number of frames per track session to minimize the occurences of tracking drift in the dataset. Although it may seem intuitive that using a higher number of frames per track is more advantageous than a lower number of frames, we qualitatively evaluate the features learned on the same unlabeled dataset using \textit{5 frames/track} (default setting in this paper) and \textit{20 frames/track} settings respectively, and show in Fig.\ \ref{fig:vis_20} that features learned in \textit{20 frames/track} setting are incredibly similar to those of \textit{5 frames/track} setting in Fig.\ \ref{fig:vis_1}. In a low frames per track setting, interest points are detected using low number of frames. If an object remains `interesting' for a long period of time, then it is very likely that multiple interest points on the same object can be detected at different timesteps, which implicitly forms a long sequence of tracked patches. Therefore, using higher number of frames per track is not significantly advantageous in this circumstance. An example of tracked patches from a track session is shown in Fig.\ \ref{fig:flow}(a).

\section{Adaptive observational model}
The observational model we use in this paper is both discriminative and adaptive. It is discriminative in the sense that it utilizes a supervised binary classifier to classify tracking observations into positive (target) class and negative (background) class. We also retrain the classifier periodically with new training samples to keep the observational model adapted to appearance changes of the target object and background over time. Observations are represented using representations learned from the stacked autoencoders introduced previously.

\subsection{Online training samples collection}
In the first frame of tracking, there are no preceding positive samples which can be concatenated with the current tracking target to form a positive training set. Therefore, in this circumstance, a set of positive samples are collected from regions that are located few pixels away from the target object. Let $ (x^t, y^t) $ denote the target object's coordinate on horizontal and vertical axes respectively in $ t $-th frame. In the first frame where $ t = 1 $, a positive sample is collected at the location $ (x^1_+, y^1_+) $ :

\begin{equation}\label{eq:6}
\mathit{j}^1_+= \mathit{j}^1+DU(-v,v)
\quad, \quad \forall \mathit{j} \in \{x,y\}
\end{equation}

\noindent
where $ v \in \mathbb{Z} $ determines the maximum pixel translation on any of the axes, $ DU(-v,v) \in \mathbb{Z} $  is a random integer sampled from a discrete uniform distribution in the range of $ [-v,v] $. In the subsequent frames where $ t > 1 $, every predicted target object observation is added to the positive sample set, by replacing the oldest positive sample in the set. Considering the fact that early target observations are more likely to be reliable predictions of the tracker, positive samples from early $ F_{es} \in \mathbb{Z} $ number of frames are kept permanently in the training set, to alleviate tracking drift. We too retain positive samples of recent $ F_{r+} \in \mathbb{Z} $ frames in the training set, in case of occasional false positive training samples the tracker obtains.

Unlike positive samples, we use only a single procedure to collect negative samples in all the frames. In this paper, a novel method to collect negative samples is proposed, in such a way that the negative samples have a little overlapping regions with the target object and they are somehow coherent with the particle filter's dynamic model in our proposed tracker. The rationale behind this coherence is that tracker which tracks fast-moving objects require some negative samples farther from the target, whereas tracker which tracks slow-moving objects does not need far negative samples. In the context of visual tracking, a particle filter's dynamic model generally requires normally-distributed translational affine parameters (refer to Section 5.1) to model the spatial translations of the target object. Let $ \sigma_x \in \mathbb{R} $ and $ \sigma_y \in \mathbb{R} $ denote the normal distribution's standard deviations for translational affine parameters on horizontal and vertical axes respectively. At $ t $-th frame, a negative sample is collected at the location $ (x^t_-,y^t_-) $ :

\begin{equation}\label{eq:7}
\mathit{j}^t_- = \mathit{j}^t+(S^t_\mathit{j} \times\varphi\times sgn(r_\mathit{j})) + r_\mathit{j}
\quad, \quad \forall \mathit{j} \in \{x,y\}
\end{equation}

\noindent
where $ S^t_\mathit{j} \in \mathbb{R} $ refers to either width $ S^t_\mathit{x} $ or height $ S^t_\mathit{y} $ of target object, $ sgn() $ denotes the sign function, $ \varphi \in \mathbb{R} $ is a constant determining the maximum overlaps between negative samples and target object on each of the axes. $  r_\mathit{j} \in \mathbb{R} $ is a random value computed as :

\begin{equation}\label{eq:8}
 r_\mathit{j} = \mathcal{N}(0,\eta\times \sigma_j)
\quad, \quad \forall \mathit{j} \in \{x,y\}
\end{equation}

\noindent
where $ \mathcal{N}(0,\eta\times \sigma_j) \in \mathbb{R} $ is a random number sampled from a normal distribution, with zero mean and $ \eta\times \sigma_j \in \mathbb{R} $ as its standard deviation. $ \eta \in \mathbb{R} $ is a constant multiplier. The second term of Eq.\ (\ref{eq:7}) specifies the maximum overlaps between negative samples and target object, whereas the third and last term of Eq.\ (\ref{eq:7}) determines the gap between negative samples and target object. At every frame, $ N_{-s} \in \mathbb{Z} $ (where $ N_{-s} \gg 1$) number of negative samples are collected and they replace the negative samples of the oldest frame in the training set. Similar to positive samples, negative samples of recent $ F_{r-} \in \mathbb{Z} $ frames are retained in the training set to alleviate visual drift. At any time of tracking, more negative samples than positive samples will be present in the training set, which causes class imbalance problem. We address this problem in Section 4.3.

\subsection{Feature extraction for visual tracking}
After training the stacked autoencoders in Section 3 offline, it is transferred for use in visual tracking. Both first and second layers of the stacked autoencoders are used to extract dense local features (amplitude and phase features) from tracking observations and training samples. Before performing feature extraction, all tracking observations are normalized to a standard tracking template size of $ 32 \times 32 $. Image size of $ 32 \times 32 $ is a good balance between computational efficiency and image details.

As in Section 3.2, local $ d_1 $-dimensional patches are densely extracted from tracking observations with $ k_1 $ stride, and passed into the first layer to obtain first-layer features. Subsequently, second-layer features are densely extracted from the first-layer feature map with a stride of $ k_2 $. Finally, the convolved representations from both the layers are concatenated to form a final representation for visual tracking. Due to the use of relatively large convolution strides, spatial pyramid pooling \citep{lazebnik2006beyond} is not performed on the convolved representations. Doing so would greatly encourage translational invariance, which is not favourable in visual tracking \citep{fan2010human}.

\subsection{Supervised binary classification}
After performing feature extraction on positive and negative samples, a training dataset with approximate labels is obtained. A linear binary classifier is adopted to distinguish between target object and background during tracking. Linear classifiers are less prone to overfitting, and high-level representations such as our learned deep representations are likely to be linearly separable. The classifier used is logistic regression due to its capability of providing predictions in probability estimates. Probability estimates or soft labels are more useful than hard labels in the case where we want to identify the most likely target object candidate among many other candidates. Taking into account the class imbalance problem in the training set, a class-weighted logistic regression is proposed for our observational model. Let $ \pmb{z}_i \in \mathbb{R}^{r \times 1} $ denote the final tracking representation (refer to Section 4.2) for $ i $-th training sample, $ Z^+ = [\pmb{z}_{1^+},\pmb{z}_{1^+}...,\pmb{z}_{D^+}] \in \mathbb{R}^{r \times D^+} $ represents the positive training set with their respective labels as $ Y^+ = [y_{1^+},y_{2^+},...,y_{D^+}]^T \in \{-1,+1\}^{D^+ \times 1}$, $ D^+ \in \mathbb{Z} $ is the number of training samples, $ r $ indicates the number of features in the final tracking representation $ z $. The negative sample counterparts of $ Z^+ $, $ Y^+ $, and $ D^+ $ are denoted as $ Z^- \in \mathbb{R}^{r \times D^-} $, $ Y^- \in \{-1,+1\}^{D^- \times 1} $, and $ D^-  \in \mathbb{Z} $ respectively. The logistic regression classifier is trained by optimizing :

\begin{equation}\label{eq:9}
\underset{\pmb{\mathfrak{w}}}{\mathrm{min}}
\: \mathcal{C}^+
\sum_{i^+ = 1}^{D^+} log(1+e^{y_{i^+}\pmb{\mathfrak{w}}^T\pmb{z}_{i^+}})
+
\mathcal{C}^-
\sum_{i^- = 1}^{D^-} log(1+e^{y_{i^-}\pmb{\mathfrak{w}}^T\pmb{z}_{i^-}})
\end{equation}

\noindent
where $ \mathcal{C}^+ \in \mathbb{R}$ is the weight parameter for positive-class logistic cost and $ \mathcal{C}^- \in \mathbb{R}$ is the weight parameter for negative-class logistic cost. To balance the learning contributions from both classes, $ \mathcal{C}^+ $  and $ \mathcal{C}^- $ are set in such a way that they are inversely proportional to $ D^+ $ and $ D^- $ respectively. Additionally, weight decay or $ L^2 $ weight regularization, $ \| \pmb{\mathfrak{w}} \|^2_2 $  is added to the cost function in Eq.\ (\ref{eq:9}), to penalize large weights and therefore reducing overfitting. In the prediction stage, the trained logistic regression classifier computes the probability or confidence score as follows:

\begin{equation}\label{eq:10}
f(z_c) = \frac{1}{1+e^{-\pmb{\mathfrak{w}}^T\pmb{z}_c}}
\end{equation}

\noindent
where $ \pmb{z}_c \in \mathbb{R}^{r \times 1}$ is the representation of a target object candidate. The candidate with the highest probability is chosen to be the target object.

Besides initializing the classifier in the first frame of tracking, the maximum probability estimate among all tracking observations is checked every $ F_f \in \mathbb{Z} $ frames, and if the maximum probability obtained is below a threshold $ \Upsilon $, then the classifier is retrained with the current training set. $ F_f $ is set to a small value to allow fast update of the classifier, in case of abrupt appearance changes to the target object. Without the probability threshold $ \Upsilon $, the update would be too often that small and trivial errors are more likely to be accumulated, eventually causing tracking drift. To account for the poorly diversified positive samples in the early frames, the maximum probability check is done every frame for early $ F_{et} \in \mathbb{Z} $ frames. Finally, there are also circumstances whereby the target object does not change by much (maximum probability remains very high) but the background has changed. To allow new background information to be learned by the classifier in a slow manner, the classifier is retrained if it has not been updated for a number of $ F_s \in \mathbb{Z} $ frames, where $ F_s \gg F_f $.

\section{Proposed tracker}

Before visual tracking can be carried out, the adaptive observational model in Section 4 should be integrated into an object state estimation method.  For this, particle filter \cite{gordon1993novel}\cite{arulampalam2002a} is chosen over other methods (e.g., Kalman filter) due to its nonlinearity, non-Gaussian assumption, and capability of maintaining multiple hypotheses. In the last part of this section, we contrast our proposed tracking method with the reviewed state-of-the-art representation learning trackers.

\subsection{Particle filter}
Particle filter is an implementation of Bayesian recursive filter. The purpose of the filter is to estimate the target state $ \hat{s_t} $ ̂ of a dynamical system, based on a sequence of observations $ x_{1:t}=\{x_1,x_2,...,x_t\} $ up to time $ t $  :

\begin{equation}\label{eq:11}
\hat{s_t} = \mathrm{argmax} \: p(s_t|x_{1:t})
\end{equation}

\noindent
The posterior distribution $ p(s_t|x_{1:t}) $ is inferred via Bayes theorem in a recursive manner :

\begin{equation}\label{eq:12}
p(s_t|x_{1:t}) = \frac{p(x_t|s_t)p(s_t|x_{1:t-1})}{p(x_t|x_{1:t-1})}
\end{equation}

\noindent
where the prior distribution $ p(s_t|x_{1:t-1}) = \int  p(s_t|s_{t-1})\\p(s_{t-1}|x_{1:t-1})  \mathrm{d}s_{t-1} $. The distribution $ p(s_t|s_{t-1}) $ expresses the state transition or dynamic model, and $ p(x_t|s_t) $ denotes the observation likelihood function tied with the observational model.

In particle filtering, the posterior distribution $ p(s_t│x_{1:t}) $ is constructed recursively using a finite set of random samples (called particles) $ \{s_t^i,i=1,2,...,N_s \} $ with importance weights $ \{\mathit{w}_t^i,i=1,2,...,N_s \} $, where $ N_s $ is the number of particles. Each particle corresponds to a hypothesis of the state. Given a candidate particle $ s_t^i $ drawn from an importance distribution $ q(s_t | s_{1:t-1},x_{1:t} ) $, the weight of the $ i $-th particle is computed as:

\begin{equation}\label{eq:13}
\mathit{w}^i_t =
\mathit{w}^i_{t-1}
\frac{p(x_t|s^i_t)p(s^i_t|s^i_{t-1})}
{q(s^i_t|s^i_{1:t-1},x_{1:t})}
\end{equation}

\noindent
In the context of visual tracking, the importance distribution $ q(s_t|s_{1:t-1},x_{1:t} ) $ is generally chosen to be $ p(s_t|s_{t-1} ) $, the dynamic model. Therefore, only the observational mode $ p(x_t|s_t ) $ has influence on the weight update of the particles.

The dynamic model $ p(s_t|s_{t-1} ) $ propagates the particles from time $ t-1 $ to $ t $, describing the temporal transition of target states between time steps.  For model-free visual tracking using rectangular bounding box, we employ affine transformation parameters (e.g., translation, scale, aspect ratio of the bounding box) as state elements to approximate the motion of target object between frames. The dynamic model is formulated such that each state element in $ s_t $ is modeled independently by a normal distribution centered at its previous state $ s_{t-1} $ :

\begin{equation}\label{eq:14}
p(s_t|s_{t-1}) =
\mathcal{N}(s_{t-1},Q)
\end{equation}

\noindent
where $ Q $ is a diagonal covariance matrix whose elements are the variances of the affine transformation parameters.

Although dynamic model plays an important role in particle filter-based trackers, the observational model $ p(x_t|s_t ) $ is the key factor to a tracker’s performance when dealing with various tracking challenges. In this paper, the observation likelihood is exponentially proportional to the confidence score $ f(\pmb{z}_t ) $ given by the periodically updated linear classifier at time $ t $ :

\begin{equation}\label{eq:15}
p(x_t|s_t )\propto\mathrm{exp}⁡(f(\pmb{z}_t ))
\end{equation}

\noindent
The exponentiation is to penalize low-weighted particles so that they are less likely to be chosen in particle resampling.

Excluding the offline training of the stacked autoencoders with temporal slowness, the high-level summary of our proposed tracker is given in Algorithm \ref{alg:1}. We refer to our proposed tracker as Deep Slow Tracker (\textbf{DST}).

\renewcommand{\algorithmicrequire}{\textbf{Input:}}
\renewcommand{\algorithmicensure}{\textbf{Output:}}
\begin{algorithm}[!htb]
\renewcommand\thealgorithm{1.}
\caption{Deep Slow Tracker}
\label{alg:1}
\begin{algorithmic}[1]
\REQUIRE tracking frames $ F_1,...,F_T $
\ENSURE target object states $ \hat{s_1},...,\hat{s_T} $
\STATE Transfer stacked autoencoders for feature extraction
\IF{$ t == 1$}
\STATE Initialize classifier with positive samples (small translations) and negative samples
\STATE Store training samples
\ELSE
\STATE Estimate $ \hat{s_t} $ using particle filter
\STATE Collect negative training samples
\STATE Store positive sample (target object at $ t $-th frame) and negative samples
\IF{$ t > F_{r+} + F_{es} $}
\STATE Remove the oldest frame's positive sample after $ F_{es} $
\ENDIF
\IF{$ t > F_{r-} + F_{es} $}
\STATE Remove the oldest frame's negative samples after $ F_{es} $
\ENDIF
\IF{max probability $ < \Upsilon $ \AND frames passed without update $ == F_f $}
\STATE Retrain linear classifier
\ELSIF{max probability $ < \Upsilon $ \AND  $ t <= F_{et} $}
\STATE Retrain linear classifier
\ELSIF{frames passed without update $ == F_s $}
\STATE Retrain linear classifier
\ENDIF
\ENDIF
\end{algorithmic}
\addtocounter{algorithm}{-1}
\end{algorithm}

\subsection{Comparison with other representation learning trackers}
Our proposed tracking method, DST is both similar to and different from the reviewed representation learning trackers \citep{wang2012transferring}, \citep{liu2013online}, \citep{jonghoon2013tracking}, and \citep{wang2013learning} in some ways. In terms of datasets used for training the representation learning models,  \citep{wang2012transferring}, \citep{jonghoon2013tracking}, \citep{wang2013learning}, and DST trains on datasets unrelated (self-taught learning \citep{raina2007self}) to the tracking video sequences. To allow the learned filters to be specific to the tracking environment, \citep{liu2013online} performs online training solely on image patches sampled from the tracking sequences itself. The datasets used in \citep{wang2013learning} and DST are generic and unlabeled, thus they are different from \citep{wang2012transferring} and \citep{jonghoon2013tracking} which use datasets with some specified object classes. However, for the sake of generality, we cannot assume objects in real-world applications share similar appearances with the limited object classes. Besides, the dataset used in DST is different from others, in the sense that we train the stacked autoencoders on tracked image patches instead of temporally uncorrelated object recognition datasets used in \citep{wang2012transferring}, \citep{liu2013online}, \citep{jonghoon2013tracking}, and \citep{wang2013learning}. All of the trackers use raw image patches to learn representations from, except \citep{wang2012transferring} which extracts SIFT features from the patches as bases to build a sparse coded dictionary.

In terms of observational model’s adaptivity, \citep{jonghoon2013tracking} is the only tracker that uses a non-adaptive offline classifier to distinguish between target object and object. \citep{wang2012transferring} and \citep{liu2013online} including DST employ linear classifiers which are independent of representation learning models, to build adaptive observational models. \citet{wang2013learning} uses a more sophisticated way for classification during tracking, by fine-tuning the deep neural network (unrolled from pre-trained stacked autoencoders) using classification error cost function. This results in a nonlinear classifier.  Supervised fine-tuning too can be applied to our proposed autoencoder but the advantages from the temporal slowness constraint might be diminished as a result of minimizing only classification errors.

\section{Experiments}
In this section, we describe the implementation details and parameter settings of DST, along with the experimental setups for the tracking experiments. DST is tested on several challenging sequences, against 7 state-of-the-art trackers. We then present the results from the experiments in quantitative and qualitative means.

\subsection{Implementation details}
In this subsection, we provide the parameter settings for the parameters described in previous section. All parameter settings of the proposed method are obtained empirically.

One of the most important aspects of DST is the offline training of the stacked autoencoders (Section 3). In terms of sequential training datasets, $ 8 \times 8 $ tracked patches are employed for training the first layer and the second layer trains on $ 14 \times 14 $ tracked patches. The number of track sessions $ N_T $ is $ 15000 $ and the number of frames per track $ N_F $ is 5. Next, the weight parameters $\alpha$ and $\gamma$ in the temporally slow autoencoder cost function (Eq.\ (\ref{eq:2}) are set as $ [100, 20] $ and $ [300, 20] $ for first and second layer respectively. To get the convolved representation for training the second layer, first layer features are densely extracted from $ 14 \times 14 $ tracked patches with spatial stride $ k_1 = 6 $. During visual tracking, first layer features are densely extracted from $ 32 \times 32 $ tracking observations with the same spatial stride of $ k_1 = 6 $, and second layer features are densely extracted from the first layer feature map with spatial stride $ k_2 = 2 $. For unconstrained optimization of the autoencoders, we employ off-the-shelf Limited-memory Broyden-Fletcher-Goldfarb-Shanno (BFGS) algorithm that is relatively memory-efficient and fast-converging \citep{ngiam2011optimization}. The optimization process stops once it reaches a fixed number of iterations, which is 200 in this paper.

In the adaptive observational model of DST, training samples of binary classes are collected online (Section 4.1) to train a linear classifier. The gap parameter $ v $ to collect first-frame positive samples is set as $ 1 $ and the constant multiplier $ \eta $ to collect negative samples is set as $ 2 $. To alleviate tracking drift, positive samples from early $ F_{es} = 15 $  frames and recent $ F_{r+} = 55 $ are retained in the current training set. Likewise, negative samples of recent $ F_{r-} = 15 $ are retained, and $ N_{-s} = 25 $ number of negative samples are collected per frame. The classifier update frequency parameters (Section 4.3) $ F_f $, $ F_et $, and $ F_s $ are set as $ 5 $, $ 10 $, and $ 25 $ respectively. As a parameter to allow quick observational model update, the probability threshold $\Upsilon $ is configured as $ 0.99 $. The above parameter settings are fixed for all benchmark sequences and we fix the number of particles in particle filter (Section 5.1)  to  $ 1000 $. Our proposed tracker, DST is implemented in MATLAB without code optimization and it runs at about $ 1.8 $ frames per second.

\begin{table}[t]
\resizebox{\linewidth}{!} {
  \centering
    \begin{tabular}{l||rrrrrrrrr}
\hline
  Sequence    & \textbf{DST} & ASLA  & CT    & DLT   & IVT   & ODFS  & PLS   & SPT & TLD \\ \hline
bird2 & {\color{red}97.0}  & {\color{blue}82.8}  & 88.9  & 46.5  & 16.2  & 61.6  & 53.5  & 16.2 & 32.3 \\
board & {\color{red}84.3} & 29.3 & 55.0 & 70.0 & 16.4 & {\color{blue}73.6} & 15.0 & {\color{blue}73.6} & 12.1 \\
bolt  & {\color{red}99.4}  & 1.4   & 0.9   & 4.0   & 1.1   & {\color{blue}6.9}  & 4.6   & 1.1 & 1.1 \\
car4  & {\color{red}100.0} & {\color{red}100.0} & 25.2  & 11.8  & {\color{red}100.0} & 25.2  & 35.2 & {\color{red}100.0} & {\color{blue}74.1} \\
cardark & {\color{red}100.0} & {\color{red}100.0} & {\color{blue}64.4} & {\color{red}100.0} & {\color{red}100.0} & 43.0  & {\color{red}100.0} & {\color{red}100.0} & 38.2 \\
cliffbar & 83.3 & 66.7 & {\color{blue}93.9} & 24.2 & 45.5 & {\color{red}95.5} & 31.8 & 53.0 & 48.5 \\
coke  & {\color{red}71.5}  & 15.1  & 23.4  & 58.1  & 52.9  & 27.5  & {\color{blue}68.0}  & 11.7 & 29.9 \\
crossing & {\color{red}100.0} & {\color{red}100.0} & 95.8  & {\color{blue}99.2}  & 48.3  & 91.7  & 25.0  & 75.8 & 33.3 \\
david & {\color{blue}95.5}  & 94.1  & 17.4  & 32.9  & 38.9  & 19.7  & 32.5  & {\color{red}95.8} & 14.9 \\
deer  & 94.4  & {\color{red}100.0} & {\color{red}100.0} & {\color{red}100.0} & {\color{red}100.0} & {\color{blue}98.6}  & 97.2  & {\color{blue}98.6} & 67.6 \\
dollar & {\color{red}100.0} & {\color{red}100.0} & {\color{red}100.0} & 37.9 & {\color{red}100.0} & {\color{blue}97.0} & 83.3 & {\color{red}100.0} & 39.4 \\
faceocc2 & {\color{red}100.0} & {\color{blue}99.8}  & 96.1  & 95.6  & 96.3  & 96.8  & 51.0  & 99.4 & 88.3 \\
football & 67.1  & 61.9  & 70.4  & 42.0  & {\color{blue}78.5}  & 74.9  & 17.1  & {\color{red}98.9} & 67.1 \\
football1 & {\color{blue}86.5}  & {\color{red}97.3}  & 5.4   & 75.7  & 83.8  & 43.2  & 33.8  & {\color{red}97.3} & 52.7 \\
jumping & 99.0  & 98.1  & 4.8   & 5.4   & {\color{blue}99.7}  & 10.5  & 9.6   & {\color{red}100.0} & 85.3 \\
mountainbike & {\color{red}100.0} & 93.4  & 26.8  & 52.2  & {\color{blue}94.3}  & 50.9  & 80.3  & 45.6 & 37.3 \\
shaking & {\color{red}100.0} & 76.2  & {\color{blue}91.0}  & 1.4   & 1.1   & 87.4  & 32.9  & 1.1 & 0.5 \\
singer1 & {\color{red}100.0} & {\color{red}100.0} & 19.9  & {\color{red}100.0} & {\color{blue}41.0}  & 19.9  & 40.2  & {\color{red}100.0} & {\color{red}100.0} \\
surfer & {\color{red}74.8} & 9.7 & 15.2 & 7.7 & {\color{blue}18.9} & 9.1 & 13.1 & 17.5 & 26.0 \\
tiger1 & 74.0 & 61.0 & {\color{blue}80.2} & {\color{red}89.3} & 9.3 & 47.7 & 28.5 & 31.9 & 61.9 \\
trellis & {\color{red}98.1}  & {\color{blue}84.4}  & 39.5  & 33.4  & 25.3  & 42.4  & 29.5  & 37.4 & 27.9 \\
walking & {\color{red}99.8}  & {\color{red}99.8}  & 54.1  & 44.4  & {\color{red}99.8}  & 49.5  & {\color{blue}79.6}  & {\color{red}99.8} & 31.1 \\
woman & {\color{red}99.6}  & {\color{blue}98.7}  & 19.8  & 92.7  & 21.1  & 20.0  & 16.9  & 20.7 & 43.1 \\ \hline
average & {\color{red}92.4} & {\color{blue}76.9} & 51.7 & 53.2 & 56.0 & 51.8 & 42.6 & 64.1 & 44.0 \\ \hline
\end{tabular}%

}
\captionsetup{justification=centering}
\caption{Average success rates (SR) in percentages. The best and second best results are presented in {\color{red}red} and {\color{blue}blue} fonts respectively.}
  \label{table:1}%
\end{table}
\begin{table}[t]
\resizebox{\linewidth}{!} {
  \centering
    \begin{tabular}{l||rrrrrrrr}
\hline
Sequence      & DST   & ASLA  & CT    & DLT   & IVT   & ODFS  & PLS   & SPT \\ \hline
bird2 & {\color{red}7.7}   & {\color{blue}10.8}  & 11.1  & 22.2  & 79.9  & 15.1  & 19.5  & 82.3 \\
board & {\color{red}6.8} & 17.5 & 11.3 & 11.8 & 30.8 & 9.6 & 42.1 & {\color{blue}7.4} \\
bolt  & {\color{red}2.7}   & 191.2 & {\color{blue}122.8} & 352.7 & 384.4 & 149.0 & 402.5 & 375.0 \\
car4  & {\color{blue}2.2}   & {\color{blue}2.2}   & 61.7  & 100.2 & {\color{red}1.9}   & 52.8  & 112.6 & 2.5 \\
cardark & 2.3   & 2.5   & 16.0  & {\color{red}1.9}   & {\color{blue}2.2}   & 33.8  & {\color{blue}2.2}   & {\color{blue}2.2} \\
cliffbar & 1.5 & 8.1 & {\color{red}1.0} & 7.6 & 7.1 & {\color{blue}1.2} & 7.6 & 6.0 \\
coke  & {\color{blue}18.7}  & 58.2  & 39.1  & {\color{red}16.8}  & 76.3  & 34.6  & 19.1  & 82.3 \\
crossing & {\color{blue}1.7}   & {\color{red}1.6}   & 3.3   & 1.9   & 2.5   & 6.3   & 138.7 & 4.4 \\
david & {\color{red}2.6}  & {\color{blue}3.4}   & 9.4   & 62.3  & 7.8   & 30.2  & 70.7  & 4.1 \\
deer  & 8.1  & {\color{red}4.5}   & 9.7   & 8.4   & {\color{blue}7.7}   & 9.1   & 8.4   & 9.1 \\
dollar & {\color{blue}1.0} & {\color{red}0.8} & 1.9 & 13.8 & 3.3 & 1.8 & 3.9 & 1.2 \\
faceocc2 & {\color{red}5.8}   & {\color{blue}8.3}   & 12.7  & 9.7   & {\color{blue}8.3}   & 9.2   & 64.2  & 11.4 \\
football & {\color{blue}6.7}   & 16.2  & 14.4  & 40.2  & 15.1  & 13.2  & 95.6  & {\color{red}4.6} \\
football1 & 7.2   & {\color{blue}6.3}   & 22.6  & 17.3  & 8.5   & 10.0  & 28.0  & {\color{red}4.5} \\
jumping & {\color{blue}4.3}   & 4.5   & 45.6  & 73.4  & 4.9   & 15.1  & 57.6  & {\color{red}4.2} \\
mountainbike & 8.3   & {\color{red}6.9}   & 188.1 & 13.9  & {\color{blue}7.2}   & 120.9 & 11.0  & 135.4 \\
shaking & {\color{red}6.8}   & 12.5  & {\color{blue}9.7}   & 37.8  & 141.9 & 10.5  & 22.1  & 100.4 \\
singer1 & 6.9   & {\color{blue}6.4}   & 12.9  & {\color{red}5.3}   & 13.6  & 11.1  & 8.0   & 6.8 \\
surfer & {\color{red}16.2} & 119.1 & 94.9 & 76.7 & 150.5 & 101.0 & {\color{blue}23.2} & 130.6 \\
tiger1 & {\color{red}5.1} & 14.7 & {\color{blue}8.8} & {\color{red}5.1} & 56.9 & 12.2 & 55.0 & 28.4 \\
trellis & {\color{red}4.0}   & {\color{blue}7.9}   & 40.8  & 78.1  & 101.7 & 38.7  & 47.0  & 62.8 \\
walking & 2.5   & 2.0   & 4.5   & 15.1  & {\color{red}1.7}   & 9.4   & 2.6   & {\color{blue}1.8} \\
woman & {\color{red}3.2}   & {\color{red}3.2}   & 110.6 & {\color{blue}4.6}   & 189.1 & 117.4 & 134.5 & 119.4 \\ \hline
average & {\color{red}5.8} & {\color{blue}22.1} & 37.1 & 42.5 & 56.7 & 35.3 & 59.8 & 51.6 \\ \hline
    \end{tabular}%
}
\captionsetup{justification=centering}
\caption{Average COL errors in pixels. The best and second best results are presented in {\color{red}red} and {\color{blue}blue} fonts respectively.}
  \label{table:2}%
\end{table}

\subsection{Experimental setups}\label{experimental_setups}
We evaluate DST on $ 24 $ challenging benchmark sequences. They are part of benchmark sequences compiled by \citet{wu2013online}, \citet{shu2011superpixel}, and \citet{babenko2009visual}. For a more comprehensive evaluation, these sequences include the various challenges in visual tracking such as fast motion, illumination variation, cluttered background, occlusion, pose variation and object deformation.

We compare DST's performances with $ 8 $ state-of-the-art trackers. The competing trackers are Adaptive Structural Local-sparse Appearance (ASLA) tracker \citep{jia2012visual}, Compressive Tracker (CT) \citep{zhang2012real}, Deep Learning Tracker (DLT) \citep{wang2013learning}, Incremental Visual Tracker (IVT) \citep{ross2008incremental}, Online Discriminative Feature Selection (ODFS) tracker \citep{zhang2013realtime}, Partial Least Squares (PLS) tracker \citep{wang2012object}, Sparse Prototypes Tracker (SPT) \citep{wang2013online}, and Tracking-Learning-Detection (TLD) \citep{zdenek2012tracking}. We run the experiments based on the codes provided by the authors. ASLA builds an efficient incremental sparse appearance model that takes into account the structural information of target object. Using random measurement matrix, CT generates compressed representation from Haar-like features and performs tracking discriminatively.  DLT is especially relevant because of its use of deep denoising autoencoders to learn a compact representation online for tracking. IVT uses a novel incremental PCA approach to generatively learn an updatable subspace representation online. ODFS performs online feature selection using weak classifiers to maximize the confidence of positive samples. On the other hand, PLS makes use of binary training samples to learn a low-dimensional discriminative subspace representation. SPT introduces sparsity and trivial templates into generative PCA subspace learning, to explicitly handle occlusion and motion blur. TLD meticulously combines detection, tracking, and learning components into a framework for long-term tracking. In contrast to other visual tracking literatures, we test DST against very recent state-of-the-art trackers, instead of earlier ones. The tracking results transcribed on the sequences can be viewed at \textit{\color{blue}{\seqsplit{http://www.youtube.com/user/DeepSlowTracker/videos}}}.

For particle filter-based trackers (ASLA, DLT, IVT, PLS, SPT, and DST), the affine parameter settings in particle filter's dynamic model are heuristically chosen according to the target object's nature in each benchmark sequence. No grid-search or deliberate optimization is done to obtain the settings. For fair comparisons, they are configured to share the same affine parameter settings and number of particles. For trackers that do not employ particle filter (CT and ODFS), we find the best setting for their object search window parameter from some possible values (in the range suggested by the authors), for each sequence. Since TLD carries out object detection densely on the image window, there is no search parameter to be tuned. Other parameters such as feature-related and training sample collection parameters are left in their default settings and fixed for all sequences, just like the way it is done for DST. Finally, all trackers are initialized with the same target object locations.

\begin{figure*}[!htb]
\centering
\includegraphics[scale=.55]{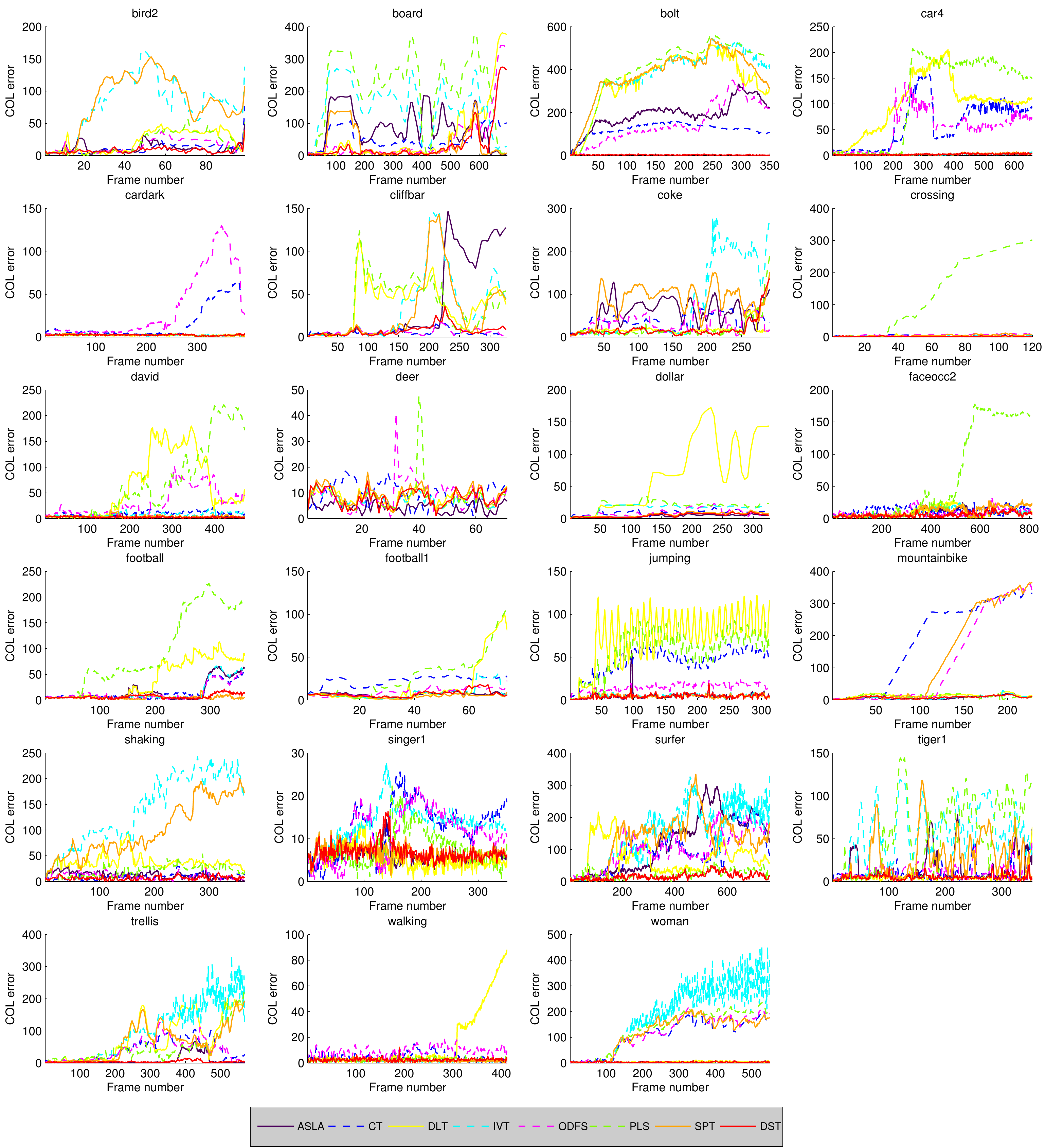}
\caption{Center-of-location (COL) error plots.}
\label{fig:col_1}
\end{figure*}

\subsection{Quantitative evaluation}
In this subsection, the trackers are evaluated quantitatively in terms of success rates (SR) and center-of-location (COL) errors. Given a ground truth bounding box $ R_G $ and a tracking result $ R_T $, SR is computed as $ \frac{area(R_T\cap R_G)}{area(R_T\cup R_G)}>0.5$. COL error is obtained by computing the Euclidean distance in pixels between the center of $ R_G $ and $ R_T $. Since all the evaluated trackers carry out random sampling, we run the trackers $ 5 $ times for each sequence and get the median results. The median results are obtained by adding max-min normalized SR and inverse COL errors from the five trials. The average SR and COL errors are shown in Table \ref{table:1} and \ref{table:2}, respectively. DST achieves the best or second best performance in most sequences, in terms of both SR and COL errors. Numerically, DST fares a lot better than its deep learning-based competitor, DLT in many of the sequences. To understand the performances of the trackers over time, the COL error plots for all tested sequences are presented in Fig.\ \ref{fig:col_1}.

\begin{figure}[!htb]
\centering
\includegraphics[scale=0.34]{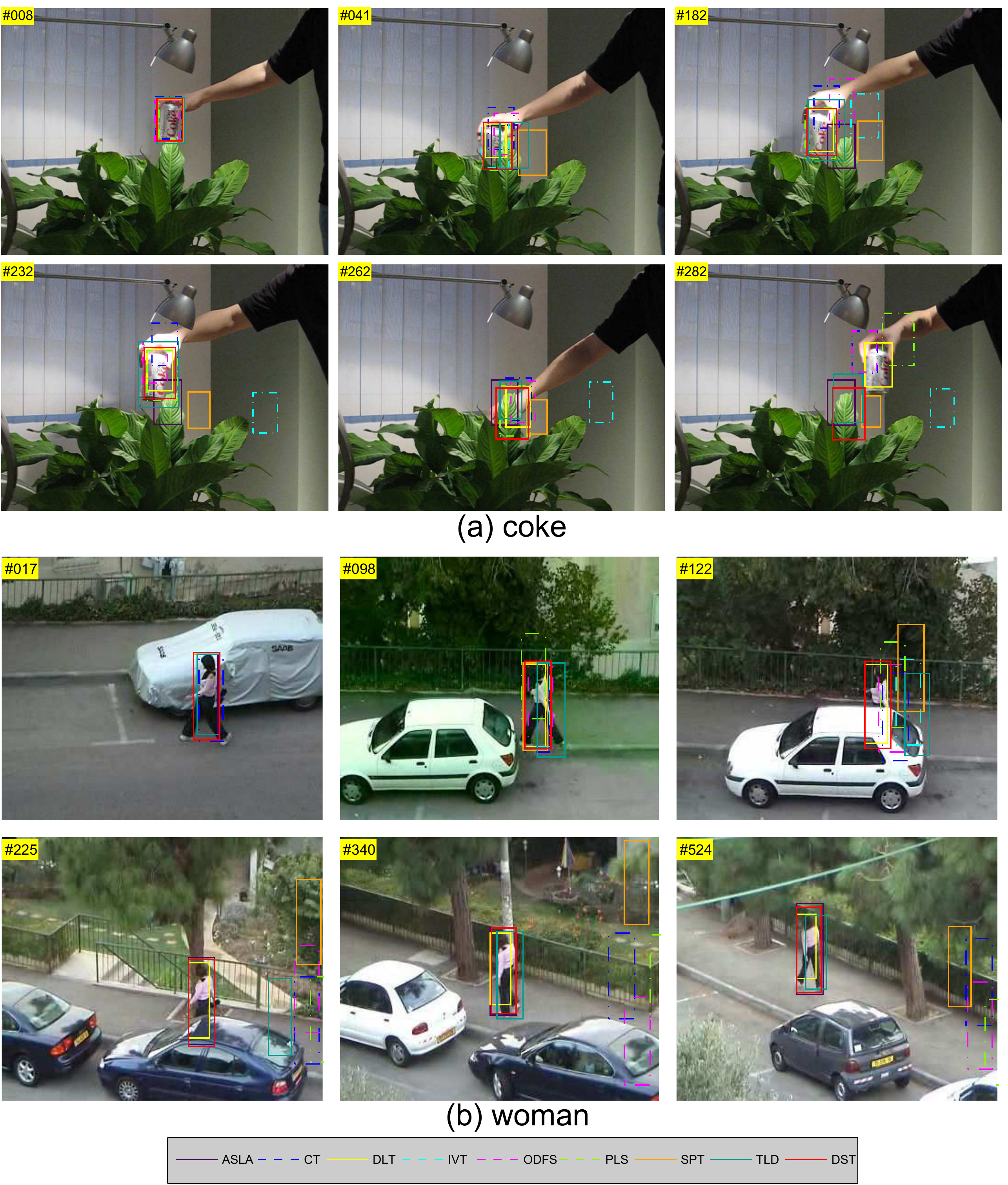}
\caption{Tracking results of sequences with occlusion: (a) \textit{coke} and (b) \textit{woman}.}
\label{fig:occ}
\end{figure}

\subsection{Qualitative evaluation}
For qualitative evaluation, we choose a number of sequences from the $ 18 $ benchmark sequences presented in Section 6.3. The sequences chosen span across the typical tracking challenges, and each of them is a representative sequence for a kind of challenge.

\textbf{Occlusion:} In the \textit{coke} sequence (Fig.\ \ref{fig:occ}(a)), the target object undergoes partial occlusion (\#041), full occlusion (\#262), out-of-plane pose change, and illumination change. ASLA, CT, ODFS, and SPT lost track of the target coming out from heavy partial occlusion (\#041). It is notable that DST performs better than DLT and as well as PLS and TLD (\#232).  IVT and SPT perform poorly because their holistic representations are not robust against partial occlusion. Our highly invariant representation learned via temporal slowness can deal well with pose change. Furthermore, the training set accumulation technique helps to alleviate post-occlusion drift. However, after the long full occlusion (\#282), only DLT can recover fully because it can retain initial target appearance model well if its particle confidence scores remain high much of the time. The target object in the \textit{woman} sequence (Fig.\ \ref{fig:occ}(b)) is a deformable woman figure heavily occluded by cars. PLS drifts away when the background is cluttered (\#098). The first occlusion (\#122) from the car is the deciding point, the trackers which make through this point can track the target well till the end. Overall, only ASLA, DLT, and DST can perform well in this sequence. Unlike other generative trackers (IVT and SPT) which are holistic, ASLA is able to handle moderate partial occlusion because of its part-based sparse representation. Even though TLD loses tracker of the object during heavy partial occlusion (\#122, \#225), its detector component can help it to recover when the object later appears unoccluded (\#340, \#524).

\begin{figure}[t]
\centering
\includegraphics[scale=0.34]{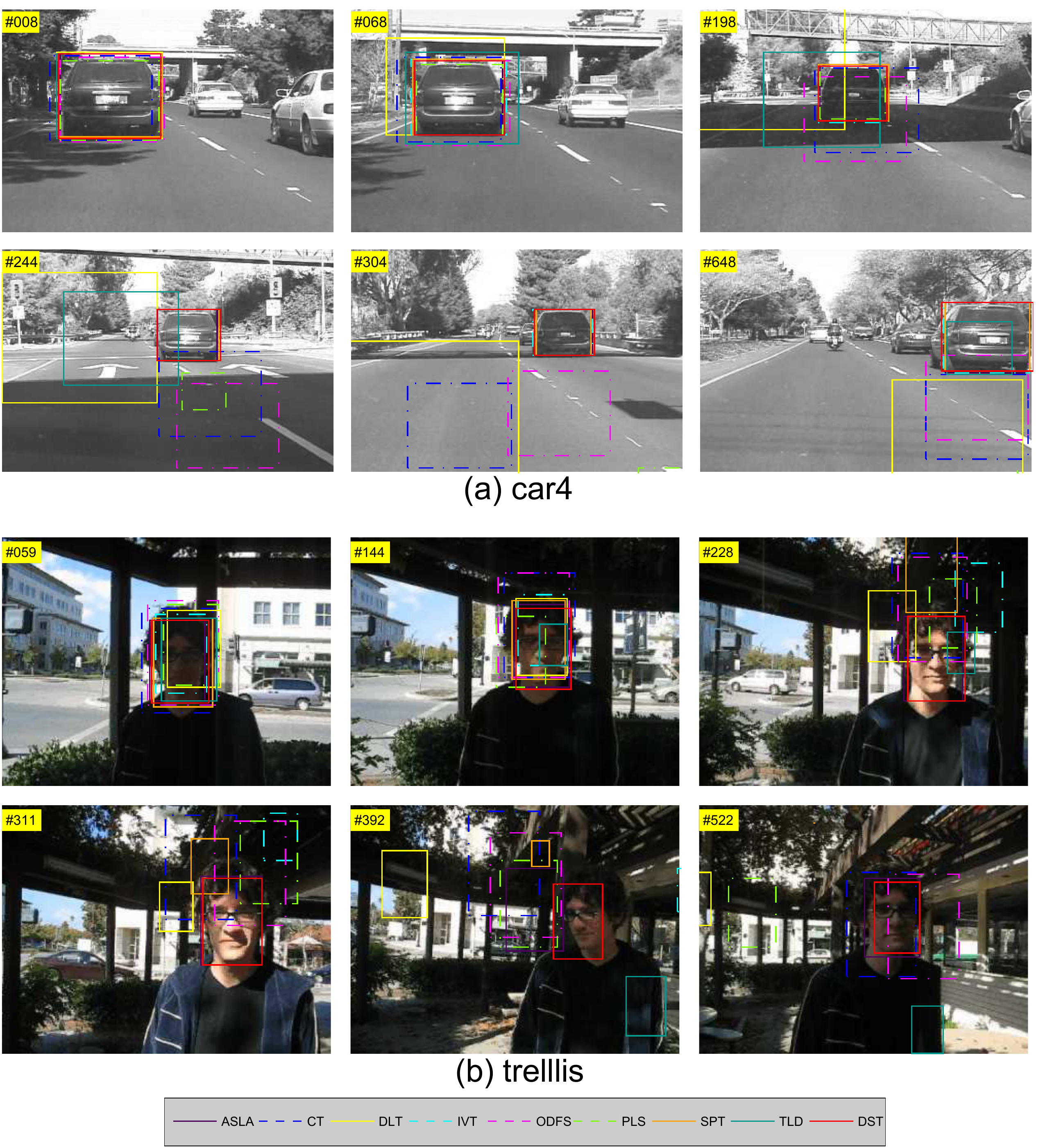}
\caption{Tracking results of sequences with illumination variation: (a) \textit{car4} and (b) \textit{trellis}.}
\label{fig:ilum}
\end{figure}

\textbf{Illumination variation:} Fig.\ \ref{fig:ilum} shows the tracking results on sequences with illumination variation. For the \textit{car4} (Fig.\ \ref{fig:ilum}(a)) sequence, the target undergoes illumination and scale changes. Although DLT employs particle filter for tracking, it does not perform well when the scale of the car changes (\#068) due to its unstable incremental training of deep neural networks. Generative representations (ASLA, IVT, SPT) perform well which can be attributed to the fact that they are robust against illumination changes (\#198, \#244, \#304). DST is on par with the generative trackers mainly due to the invariant features extracted out of background training samples, allowing relatively unchanged background candidates to be rejected even though the target's appearance has changed. In the \textit{trellis} (Fig.\ \ref{fig:ilum}(b)) sequence, there are pose changes as well as frequent illumination changes on the target. All trackers except ASLA and DST drift away from the target experience illumination change from the sunlight (\#228). Besides illumination change, DST is the only tracker which can effectively deal with out-of-plane pose change (\#392). Our novel online negative sampling method collects diverse and relevant background samples, to alleviate the problem of tracker drifting to background regions similar to the target.

\begin{figure}[t]
\centering
\includegraphics[scale=0.34]{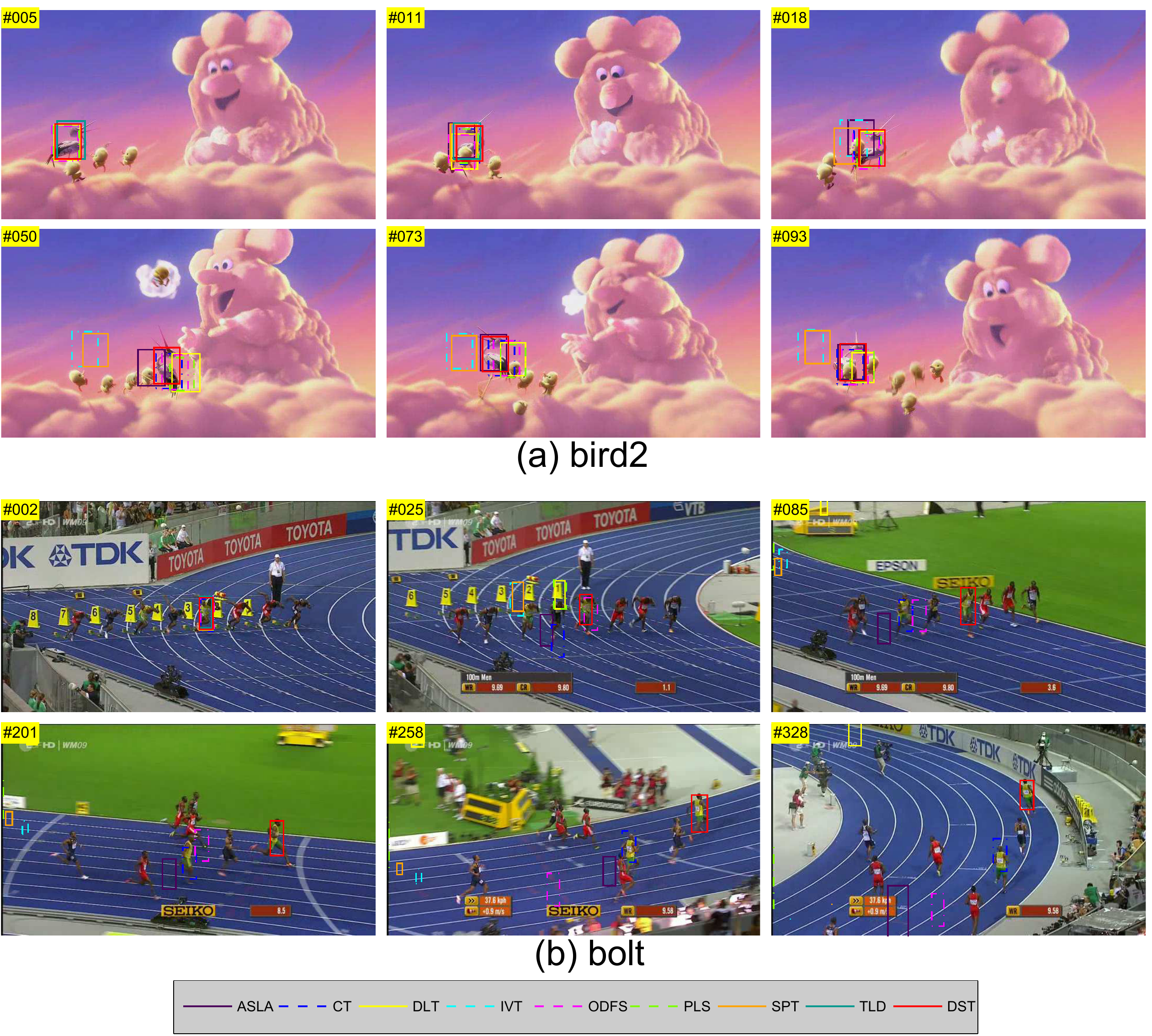}
\caption{Tracking results of sequences with pose variation and shape deformation: (a) \textit{bird2} and (b) \textit{bolt}.}
\label{fig:pose}
\end{figure}

\textbf{Pose variation and deformation:} Fig.\ \ref{fig:pose}(a) shows the tracking results of the \textit{bird2} sequence, in which the target is a bird walks back and forth while undergoing  pose variation (\#050) and partial occlusion (\#011, \#093). Other than that, the target is a deformable object which requires rectangular tracking bounding boxes to include much of background region. Generative trackers (ASLA, IVT, SPT) fail to track after heavy partial occlusion (\#018) by objects with similar appearances. Only CT and DST remain accurate after the object undergoes a large out-of-plane rotation (\#050). ASLA recovers when certain parts of the target become more recognizable with regards to its early appearance. PLS uses a non-adaptive appearance model in its second particle filtering step, which causes drift when significant appearance change such as pose change occurs. The discriminative nature of our observational model is more robust against partial occlusion by similar objects, whereas the invariant amplitude features learned by the stacked autoencoders help the tracker to deal with pose changes. In the \textit{bolt} sequence, the target object is an athlete who sprints on a track, undergoing shape deformation and gradual pose variation. \textit{bolt} (Fig.\ \ref{fig:pose}(b)) sequence is the most challenging among all tested sequences because all except DST drift away from the target at the beginning (\#025). DST performs the best in this sequence and tracks the target well until the end. The success of DST in this sequence is attributed to the highly descriptive representation, formed by the convolutionally trained second-layer features and the edge detector-like first-layer features.

\begin{figure}[t]
\centering
\includegraphics[scale=0.34]{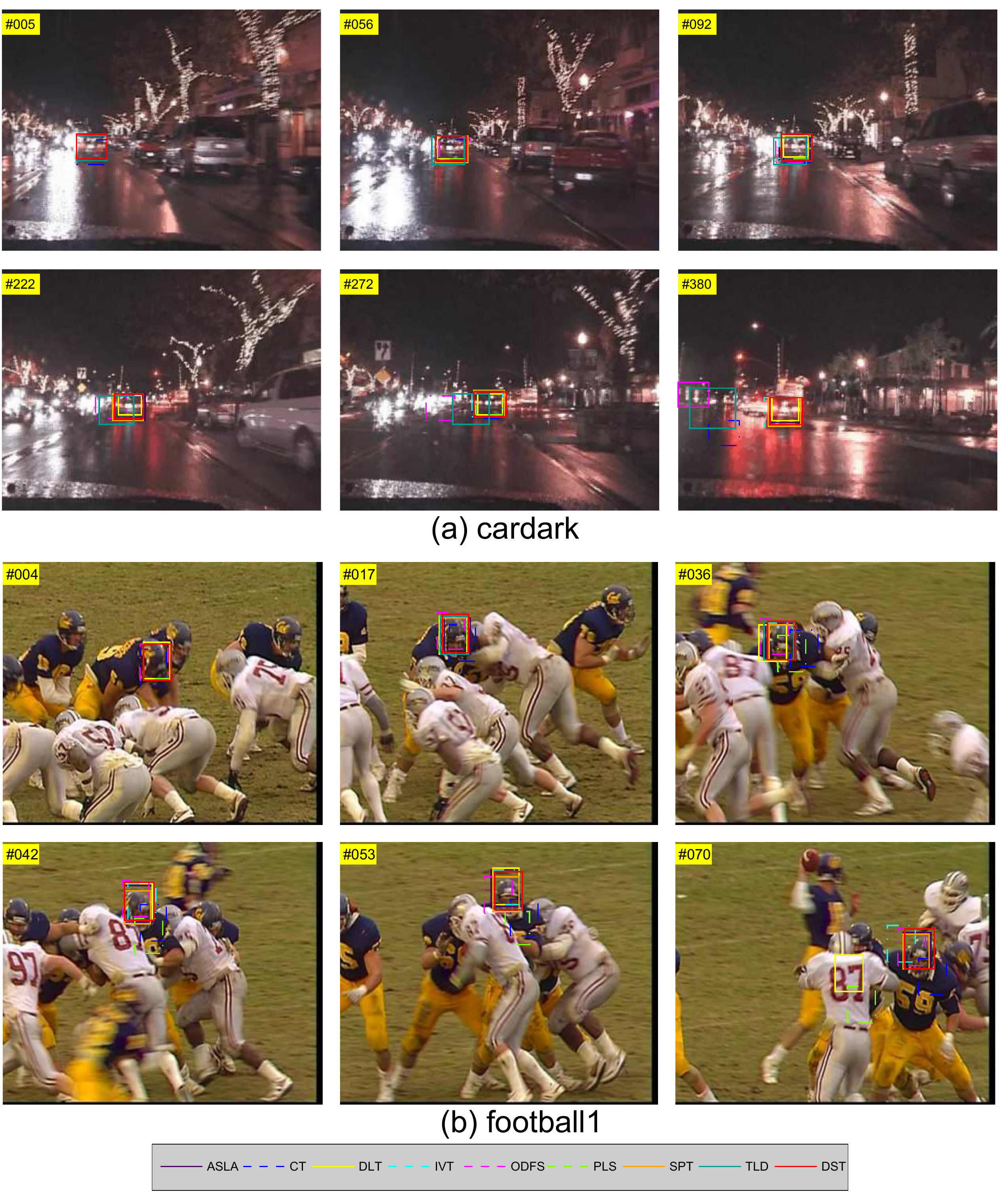}
\caption{Tracking results of sequences with cluttered background: (a) \textit{cardark} and (b) \textit{football1}.}
\label{fig:clutter}
\end{figure}

\textbf{Cluttered background:} In the \textit{cardark} sequence (Fig.\ \ref{fig:clutter}(a)), there is a low contrast between background and foreground (\#272) as well as illumination changes. CT and ODFS fail because they make use of illumination-sensitive Haar-like features as base features and perform update every frame which accumulates tracking errors easily. TLD's optical flow-based tracking causes gradual drift in the highly cluttered environment while not being able to revert to object detection mode. In situations where appearances of target and background do not change much, DLT and DST perform well because they are both updated in a slower manner depending on the maximum confidence score of all tracking observations. Fig.\ \ref{fig:clutter}(b) shows some representative tracking results of the \textit{football1} sequence, in which the target object is the helmeted head of a football athelete. The sequence is tough because the background is cluttered by atheletes with similar appearances, and the athete undergoes motion blur (\#042) and significant pose change (\#070). Overall, only ASLA, SPT, and DST perform well. ASLA's novel alignment pooling makes it less prone to drifting problem when other similar objects are around. By handling motion blur explicity, SPT avoids bad updates and performs better than IVT. DST is able to achieve similar result, using accumulation of training samples to weaken the contribution of bad training samples.

\begin{figure}[t]
\centering
\includegraphics[scale=0.34]{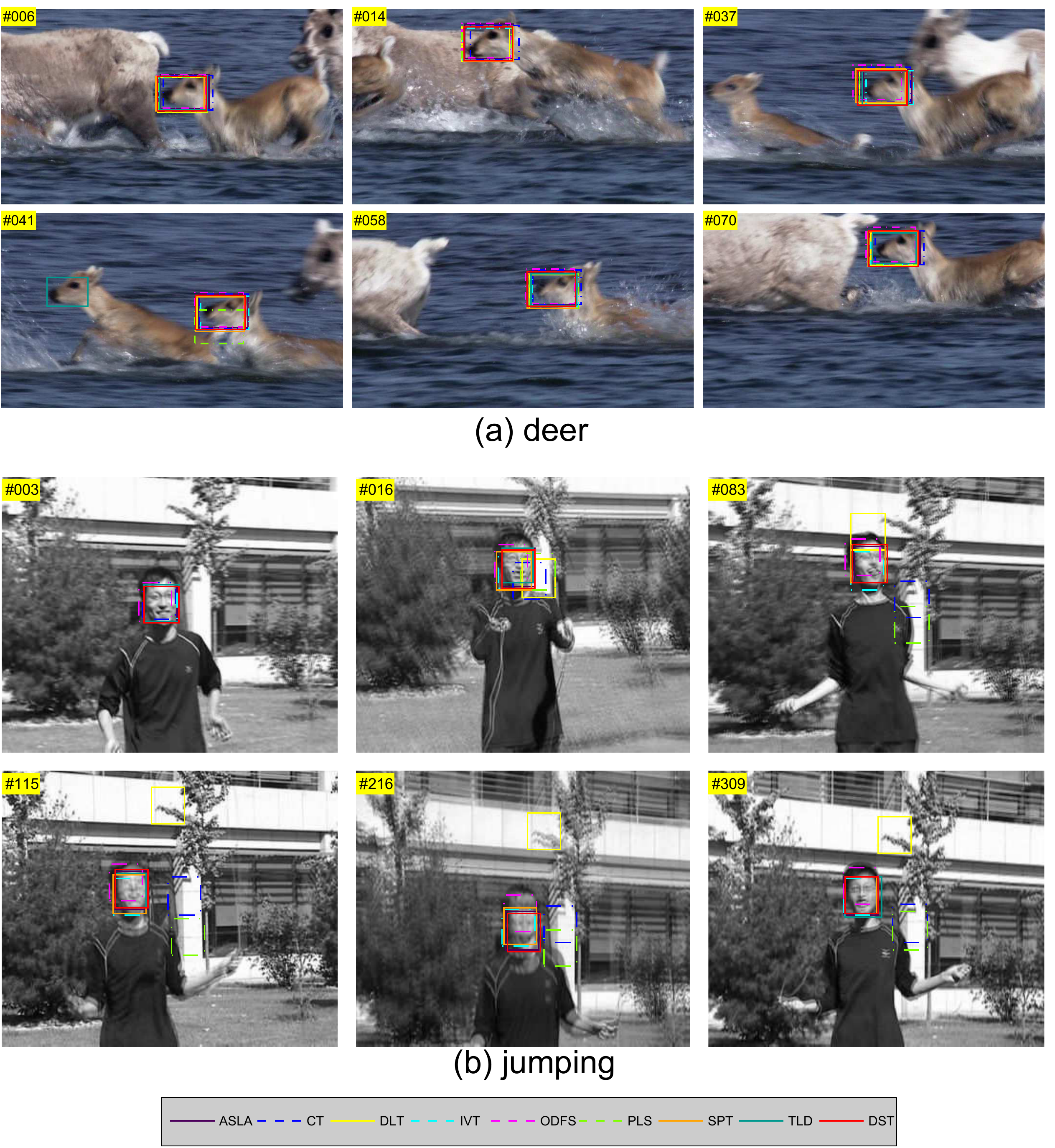}
\caption{Tracking results of sequences with fast motion and motion blur: (a) \textit{deer} and (b) \textit{jumping}.}
\label{fig:fast}
\end{figure}

\textbf{Fast motion and motion blur:} Fig.\ \ref{fig:fast} shows the tracking results on sequences with fast motion. Objects with fast motion tend to generate motion blur in images captured by the common cameras. The target object in the \textit{deer} sequence is the head of a quickly moving deer with abrupt location change and motion blur. Overall, all trackers are able to track the target due to heuristically chosen affine settings and fine-tuned search window settings. Furthermore, there is always a big contrast between the target and background (water). PLS experiences slight drift (\#041) which is attributed by the fact that its static appearance model does not consider appearance variations in the target and background over time. TLD drifts to a confusingly similar background object (\#041) when it cannot track realiably and has to switch to detection mode. For the \textit{jumping} sequence, the target object undergoes large translation between frames and exhibits significant motion blur. Nearly all discriminative trackers (CT, DLT, ODFS, PLS) except DST perform poorly because they rely on negative sampling methods that do not consider target's motion pattern. TLD works reasonably well by learning object appearance cautiously through the estimation of false negatives and false positives. Without considering negative templates, generative trackers (ASLA, IVT, SPT) track the target well. DST succeeds in this sequence by virtue of a novel negative sampling method that adapts to the particle filter's dynamic model.

\begin{table}[!htb]
\resizebox{\linewidth}{!} {
  \centering
    \begin{tabular}{l||rr|rr|rr|rr}
\hline
    Sequence & \multicolumn{2}{c|}{\textbf{DST}}    & \multicolumn{2}{c|}{HOG}   & \multicolumn{2}{c|}{LBP}  &   \multicolumn{2}{c}{SIFT}  \\ \cline{2-9}
& SR & COL & SR & COL & SR & COL & SR & COL \\ \hline

bird2 & {\color{red}97.0}  & {\color{red}7.7}   & 90.9  & 11.3  & 84.8  & 12.0  & 88.9  & 11.0 \\
board & 84.3 & 6.8 & 87.1 & 7.0 & {\color{red}95.7} & {\color{red}2.7} & 81.4 & 7.9 \\
bolt  & {\color{red}99.4}  & {\color{red}2.7}   & 1.4   & 411.3 & 52.0  & 4.9   & 89.4  & 4.0 \\
car4  & {\color{red}100.0} & 2.2   & {\color{red}100.0} & {\color{red}2.0}  & {\color{red}100.0} & 2.4   & {\color{red}100.0} & {\color{red}2.0} \\
cardark & {\color{red}100.0} & 2.3   & {\color{red}100.0} & {\color{red}2.2}   & {\color{red}100.0} & {\color{red}2.2}   & {\color{red}100.0} & 2.3 \\
cliffbar & 83.3 & 1.5 & {\color{red}97.0} & {\color{red}1.4} & 45.5 & 4.5 & 87.9 & 1.5 \\
coke  & 71.5  & 18.7  & {\color{red}86.9}  & {\color{red}14.6}  & 86.6  & 15.5  & 75.3  & 20.6 \\
crossing & {\color{red}100.0} & 1.7   & {\color{red}100.0} & 1.8   & {\color{red}100.0} & 1.7   & {\color{red}100.0} & {\color{red}1.5} \\
david & 95.5  & 2.6   & 97.5  & {\color{red}2.5}   & 99.2  & 3.7   & {\color{red}100.0} & 3.0 \\
deer  & 94.4  & 8.1   & {\color{red}100.0} & {\color{red}6.4}   & {\color{red}100.0} & 7.7   & 94.4  & 7.4 \\
dollar & {\color{red}100.0} & {\color{red}1.0} & {\color{red}100.0} & 1.1 & {\color{red}100.0} & 1.2 & {\color{red}100.0} & 1.1 \\
faceocc2 & {\color{red}100.0} & {\color{red}5.8}   & {\color{red}100.0} & 5.9   & {\color{red}100.0} & 6.0   & {\color{red}100.0} & 6.4 \\
football & 67.1  & 6.7   & 78.2  & 14.4  & 83.4  & 6.2   & {\color{red}88.4}  & {\color{red}6.0} \\
football1 & {\color{red}86.5}  & {\color{red}7.2}   & 75.7  & 12.9  & 81.1  & 10.6  & 73.0  & 24.4 \\
jumping & 99.0  & 4.3   & 99.4  & 3.7   & {\color{red}100.0} & {\color{red}4.0}   & 97.4  & 4.5 \\
mountainbike & {\color{red}100.0} & {\color{red}8.3}   & 96.1  & 8.5   & 97.8  & 8.7   & 96.9  & 9.3 \\
shaking & {\color{red}100.0} & {\color{red}6.8}   & 95.1  & 9.1   & 93.2  & 8.0   & 87.9  & 9.8 \\
singer1 & {\color{red}100.0} & 6.9   & {\color{red}100.0} & {\color{red}5.9}   & {\color{red}100.0} & 7.0   & 98.3  & 6.2 \\
surfer & {\color{red}74.8} & {\color{red}16.2} & 5.9 & 112.6 & 5.8 & 123.9 & 6.0 & 194.7 \\
tiger1 & 74.0 & 5.1 & 92.1 & {\color{red}3.3} & 92.9 & 4.9 & {\color{red}93.5} & 5.8 \\
trellis & 98.1  & 4.0   & {\color{red}99.5}  & {\color{red}3.4}   & 98.1  & 4.2   & 98.2  & 3.4 \\
walking & {\color{red}99.8}  & 2.5   & 96.4  & {\color{red}2.1}   & 89.3  & 2.2   & {\color{red}99.8}  & 1.9 \\
woman & {\color{red}99.6}  & {\color{red}3.2}   & {\color{red}99.6}  & {\color{red}3.2}   & 98.9  & 3.7   & {\color{red}99.6}  & 3.3 \\ \hline
average & {\color{red}92.4} & {\color{red}5.8} & 86.9 & 28.1 & 87.1 & 10.8 & 89.4 & 14.7 \\ \hline
    \end{tabular}%
}
\captionsetup{justification=centering}
\caption{Average SR and COL errors for hand-crafted representation trackers and DST. The best results are presented in {\color{red}red} font.}
  \label{table:3}%
\end{table}

\subsection{Discussions}
\textbf{Temporally slow representation versus hand-crafted representation:} Traditionally, visual tracking research relies on hand-crafted representations such as SIFT, HOG, and LBP. The main contribution in this paper is to take an alternative approach of learning features for visual tracking. We learn slow invariant features offline via stacked autoencoders and transfer them for online visual tacking. To this end, we isolate the merits of our proprosed tracking representation by substituting it with dense SIFT, HOG, and LBP local descriptors. All other components of our proprosed tracker remain the same. To obtain the hand-crafted descriptors for tracking, we use a reputable computer vision library package known as VLFeat \citep{vedaldi2010vlfeat}. For SIFT, we use the same spatial stride as our first layer autoencoder and find the best width (in pixels) of SIFT spatial bins, from some possible values close to the first layer's input size. Then, for HOG and LBP, we find the best setting in terms of cell sizes. Only the settings that yield the best overall results are selected for evaluation. All other parameters in the hand-crafted feature extractions are left in their default settings. Experimentally, we evaluate the handcrafted representation trackers and our proprosed tracker DST on all the $ 18 $ benchmark sequences (Section 6.2). The average SR and COL errors for each sequence are shown in Table \ref{table:3}. From the table, it is shown that DST is comparable with trackers substituted with hand-crafted representations. DST is most the well-rounded tracker, achieving the best results in terms of average SR and COL error. The advantage of the proposed representation over hand-crafted representations is especially evident on \textit{bird2}, \textit{bolt}, \textit{football1}, \textit{shaking}, and \textit{surfer} sequences, where large pose changes are prevalent.

\begin{figure}[t]
\centering
\includegraphics[scale=0.55]{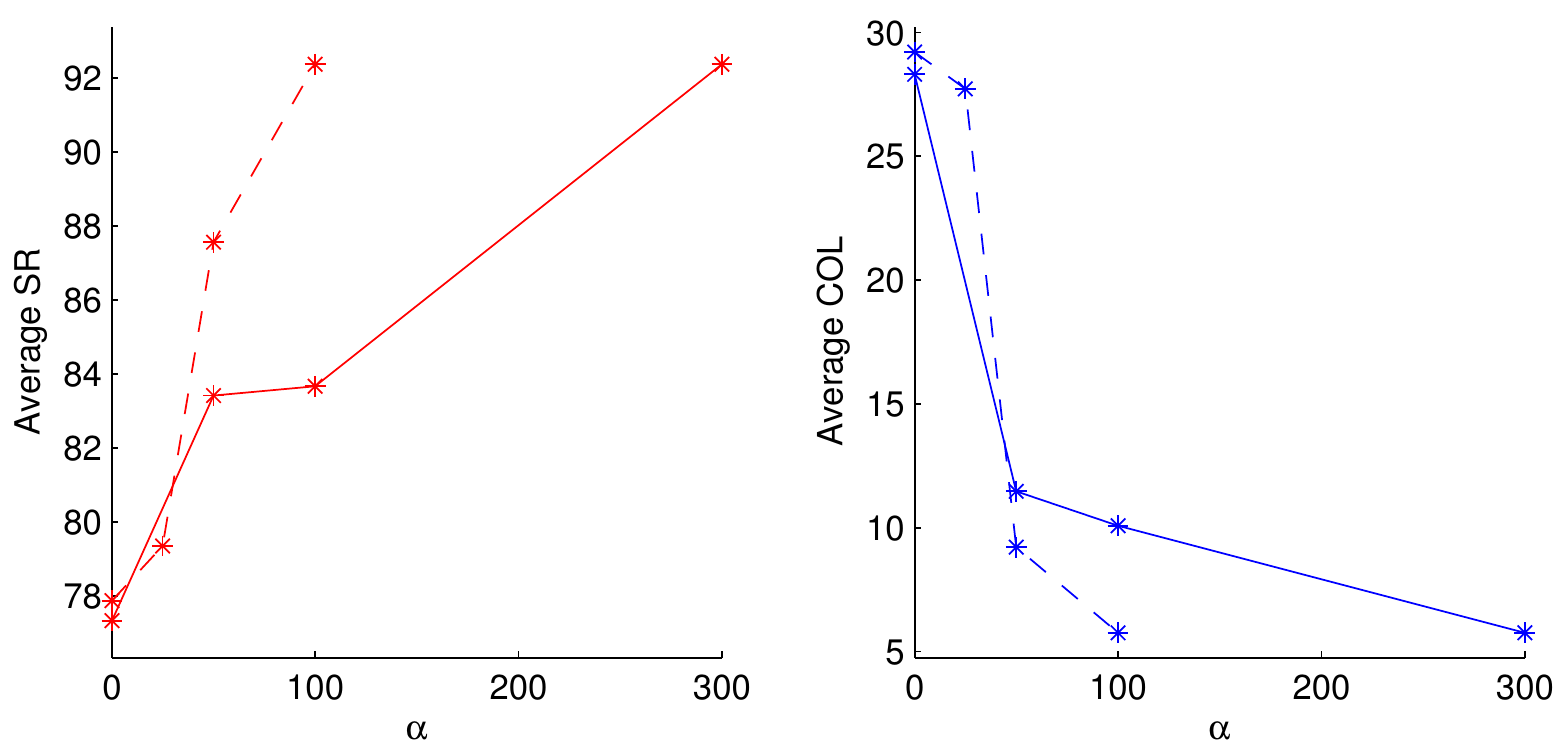}
\caption{SR (left) and COL errors (right) averaged from all sequences, with varying temporal slowness strengths $ \alpha $ of first layer (dotted line) and second layer (solid line).}
\label{fig:sr_col_1}
\end{figure}

\begin{figure}[!htb]
\centering
\includegraphics[scale=0.37]{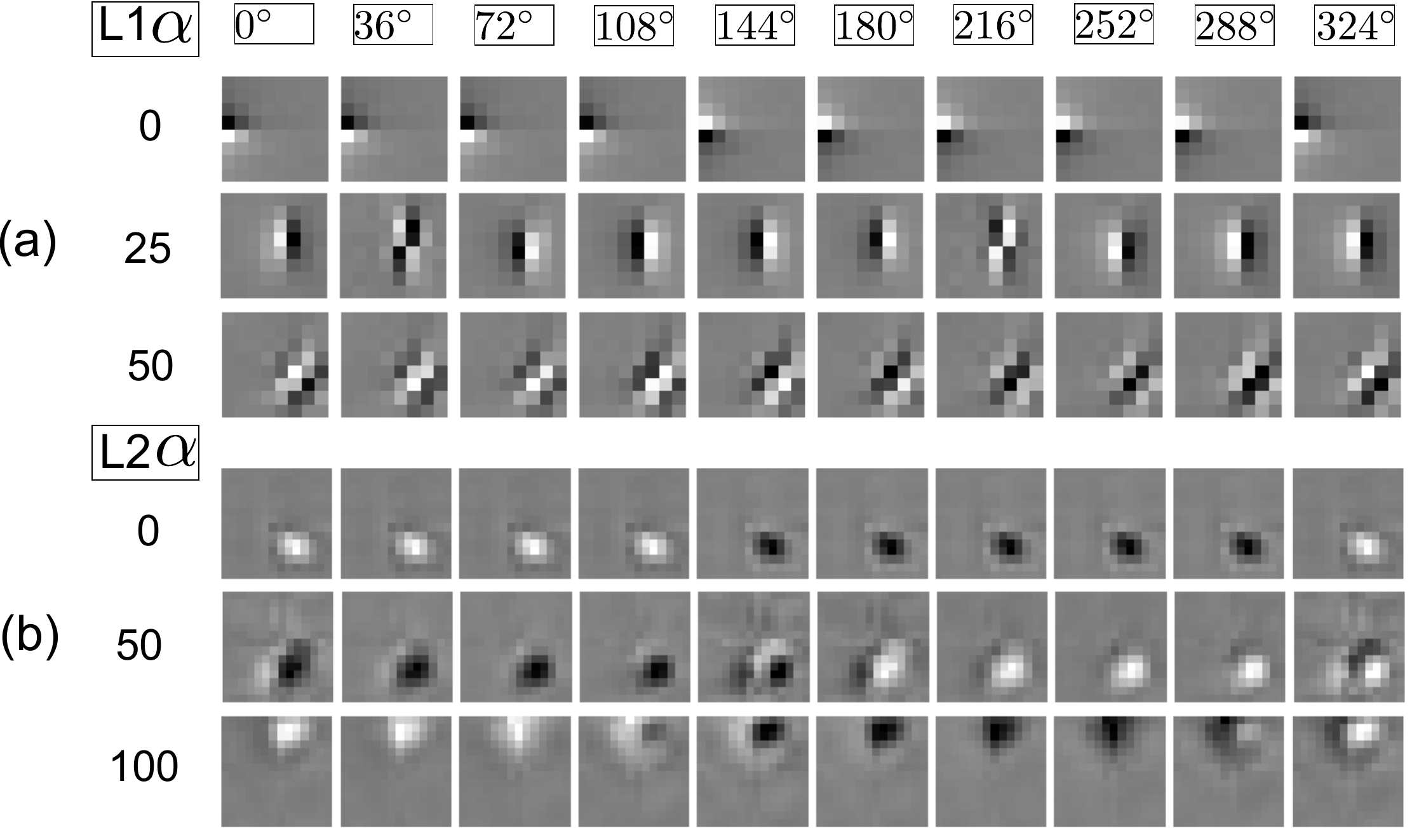}
\caption{Visualized invariances of (a) first layer and (b) second layer, with varying temporal slowness weights $\alpha $ on each row.}
\label{fig:vis_2}
\end{figure}

\textbf{Effect of temporal slowness strength:} The deep invariant representation proposed in this paper is learned by enforcing strong temporal slowness constraint Eq.\ (\ref{eq:2}) in the first and second layer autoencoders. To understand the importance of temporal slowness for visual tracking, we train the stacked autoencoders with varying temporal slowness weight $ \alpha $ and evaluate the learned representations on visual tracking tasks. The two autoencoder layers are assessed independently by fixing the temporal slowness weight of one layer and varying the slowness weight of another. The default settings for the autoencoder parameters follow the same settings in DST (Section 6.1). For the varying temporal slowness strengths, they are obtained by choosing some relevant values lower than the default parameter settings. Due to the fixing of sparsity and reconstruction weights, the reduction of temporal slowness forces the autoencoders to rely more on sparsity and reconstruction to learn features. Experimentally, the tracking representation of DST is substituted with the representations with varying temporal slowness strengths and they are evaluated on all benchmark sequences. The plots of SR and COL errors (averaged from all sequences) against the varying temporal slowness strengths for first and second layer are presented in Fig.\ \ref{fig:sr_col_1}. The plots demonstrate that tracking performance improves with increased temporal slowness strength in either layer of the proposed stacked autoencoders. To understand the effects of varying temporal slowness strengths on the  learned features visually, we present some phase-shifted optimal stimuli with varying $ \alpha $ in Fig.\ \ref{fig:vis_2}. It is worth noting that a lower temporal slowness strength produces features which are invariant to very limited transformations, and the transitions between the transformations are sudden and unsmooth. Conversely, it is the other way round for features learned with higher temporal slowness strengths, such as Fig.\ \ref{fig:vis_1}, and as well as when first layer's $\alpha$ is $ 50 $ and second layer's $ \alpha$ is $ 100 $.

\begin{figure}[t]
\centering
\includegraphics[scale=0.34]{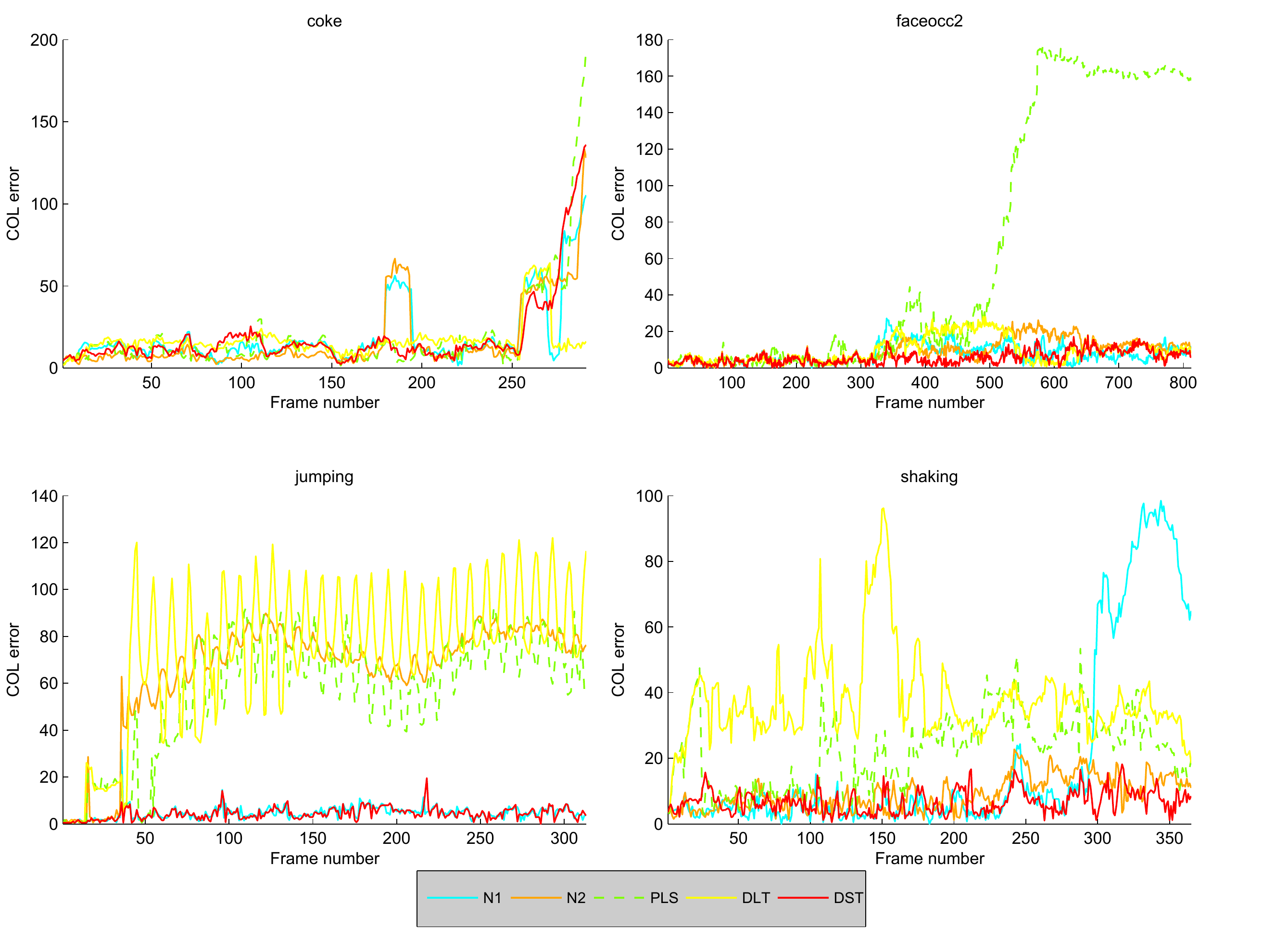}
\caption{Center-of-location (COL) error plots for negative sampling experiments.}
\label{fig:col_2}
\end{figure}

\textbf{Online negative sampling:} Online negative sampling is one of the most important components in discriminative tracking. We propose a novel negative sampling method (Section 4.1) which takes into account the maximum overlap between target object and negative samples. Moreover, the negative sampling method is adapted to the motion-pattern of the target objects, according to the affine parameter settings of particle filter. In this part, we compare our proposed negative sampling method with existing methods employed by PLS and DLT. PLS draws negative samples in an annular region \citep{wang2012object} defined by inner and outer radius, where the inner radius is the radius of circle minimally enclosing the target and outer radius is a fixed parameter. In DLT,  negative samples are collected at locations sampled from zero-mean normal distribution, in which the standard deviation is a fixed parameter multiplied by the height or width (depending on the axes) of target object. Generally, PLS collects further negative samples without overlap and DLT collects overlapping negative samples which are very near. To have a fair assessment, we substitute our negative sampling method with the methods from PLS and DLT, and use the same parameter settings. We name our tracker with PLS negative sampling method as N1 and the one with DLT negative sampling method as N2. Experimentally, we evaluate N1, N2, our proposed tracker (DST) on \textit{coke}, \textit{faceocc2}, \textit{jumping}, and \textit{shaking} challenging benchmark sequences. The COL error plots are presented in Fig.\ \ref{fig:col_2}. For comparisons, we include the tracking results from PLS and DLT trackers. Noticeably, our proposed negative sampling method is superior to N2 and comparable to N1 in overall. DST performs better than both N1 and N2 in \textit{faceocc2} in which there are abrupt lighting and appearance changes. For sequence with abrupt appearance changes and fast motion (\textit{jumping}), DST does better than N2 and it is comparable to N1. Even without the proposed negative sampling method, N1 performs better than its original counterparts (PLS) in most of the sequences (\textit{faceocc2}, \textit{jumping}, and \textit{shaking}), whereas N2 performs significantly better than DST in sequences with abrupt appearance changes (\textit{jumping} and \textit{shaking}), due to the effective deep slow representation.

\section{Conclusion}
This work exploits the temporal slowness principle to learn invariant representations for visual tracking. Temporal slowness constraint is incorporated to a autoencoder algorithm to facilitate representation learning. To allow the learned representations to be specific to visual tracking tasks, large number of tracked image patches are collected via an existing tracker to be the training set. A deep learning model is formed by stacking the autoencoders to learn higher-level invariances with temporal slowness. We then transfer the offline learned representations to an observational model for online visual tracking. The adaptive observational is formulated such that it collects more relevant negative samples and it utilizes accumulative training set to alleviate tracking drift. Tracking is carried out in the particle filter framework to estimate the target state sequentially. Compared to several state-of-the-art trackers, the proposed tracker demonstrates favourable performances in challenging benchmark sequences with various tracking challenges. In future work, we will explore the possibility of online learning of representations using temporal slowness for visual tracking, without relying on self-taught learning paradigm \citep{raina2007self}.


\bibliography{refs}

\begin{thebibliography}{51}
\providecommand{\natexlab}[1]{#1}
\providecommand{\url}[1]{\texttt{#1}}
\providecommand{\urlprefix}{URL }
\expandafter\ifx\csname urlstyle\endcsname\relax
  \providecommand{\doi}[1]{doi:\discretionary{}{}{}#1}\else
  \providecommand{\doi}[1]{doi:\discretionary{}{}{}\begingroup
  \urlstyle{rm}\url{#1}\endgroup}\fi
\providecommand{\bibinfo}[2]{#2}

\bibitem[{Li et~al.(2013)Li, Hu, Shen, Zhang, Dick, and Hengel}]{li2013asurvey}
\bibinfo{author}{X.~Li}, \bibinfo{author}{W.~Hu}, \bibinfo{author}{C.~Shen},
  \bibinfo{author}{Z.~Zhang}, \bibinfo{author}{A.~Dick},
  \bibinfo{author}{A.~V.~D. Hengel}, \bibinfo{title}{A Survey of Appearance
  Models in Visual Object Tracking}, \bibinfo{journal}{ACM Transactions on
  Intelligent Systems and Technology} \bibinfo{volume}{4}~(\bibinfo{number}{4})
  (\bibinfo{year}{2013}) \bibinfo{pages}{58:1--58:48}.

\bibitem[{Raina et~al.(2007)Raina, Battle, Lee, Packer, and Ng}]{raina2007self}
\bibinfo{author}{R.~Raina}, \bibinfo{author}{A.~Battle},
  \bibinfo{author}{H.~Lee}, \bibinfo{author}{B.~Packer}, \bibinfo{author}{A.~Y.
  Ng}, \bibinfo{title}{Self-taught learning: transfer learning from unlabeled
  data}, in: \bibinfo{booktitle}{Proceedings of International Conference on
  Machine Learning}, \bibinfo{pages}{759--766}, \bibinfo{year}{2007}.

\bibitem[{Ross et~al.(2008)Ross, Lim, Lin, and Yang}]{ross2008incremental}
\bibinfo{author}{D.~A. Ross}, \bibinfo{author}{J.~Lim}, \bibinfo{author}{R.-S.
  Lin}, \bibinfo{author}{M.-H. Yang}, \bibinfo{title}{Incremental learning for
  robust visual tracking}, \bibinfo{journal}{International Journal of Computer
  Vision} \bibinfo{volume}{77}~(\bibinfo{number}{1-3}) (\bibinfo{year}{2008})
  \bibinfo{pages}{125--141}.

\bibitem[{Liwicki et~al.(2012)Liwicki, Zafeiriou, Tzimiropoulos, and
  Pantic}]{liwicki2012efficient}
\bibinfo{author}{S.~Liwicki}, \bibinfo{author}{S.~Zafeiriou},
  \bibinfo{author}{G.~Tzimiropoulos}, \bibinfo{author}{M.~Pantic},
  \bibinfo{title}{Efficient Online Subspace Learning With an Indefinite Kernel
  for Visual Tracking and Recognition}, \bibinfo{journal}{IEEE Transactions on
  Neural Networks and Learning Systems}
  \bibinfo{volume}{23}~(\bibinfo{number}{10}) (\bibinfo{year}{2012})
  \bibinfo{pages}{1624--1636}.

\bibitem[{Kim(2012)}]{kim2012correlation}
\bibinfo{author}{M.~Kim}, \bibinfo{title}{Correlation-based incremental visual
  tracking}, \bibinfo{journal}{Pattern Recognition}
  \bibinfo{volume}{45}~(\bibinfo{number}{3}) (\bibinfo{year}{2012})
  \bibinfo{pages}{1050 -- 1060}.

\bibitem[{Jepson et~al.(2003)Jepson, Fleet, and El-Maraghi}]{jepson2003robust}
\bibinfo{author}{A.~D. Jepson}, \bibinfo{author}{D.~J. Fleet},
  \bibinfo{author}{T.~F. El-Maraghi}, \bibinfo{title}{Robust online appearance
  models for visual tracking}, \bibinfo{journal}{IEEE Transactions on Pattern
  Analysis and Machine Intelligence}
  \bibinfo{volume}{25}~(\bibinfo{number}{10}) (\bibinfo{year}{2003})
  \bibinfo{pages}{1296--1311}.

\bibitem[{Wang et~al.(2007)Wang, Suter, Schindler, and Shen}]{wang2007adaptive}
\bibinfo{author}{H.~Wang}, \bibinfo{author}{D.~Suter},
  \bibinfo{author}{K.~Schindler}, \bibinfo{author}{C.~Shen},
  \bibinfo{title}{Adaptive object tracking based on an effective appearance
  filter}, \bibinfo{journal}{IEEE Transactions on Pattern Analysis and Machine
  Intelligence} \bibinfo{volume}{29}~(\bibinfo{number}{9})
  (\bibinfo{year}{2007}) \bibinfo{pages}{1661--1667}.

\bibitem[{Collins et~al.(2005)Collins, Liu, and Leordeanu}]{collins2005online}
\bibinfo{author}{R.~T. Collins}, \bibinfo{author}{Y.~Liu},
  \bibinfo{author}{M.~Leordeanu}, \bibinfo{title}{Online selection of
  discriminative tracking features}, \bibinfo{journal}{IEEE Transactions on
  Pattern Analysis and Machine Intelligence}
  \bibinfo{volume}{27}~(\bibinfo{number}{10}) (\bibinfo{year}{2005})
  \bibinfo{pages}{1631--1643}.

\bibitem[{Grabner et~al.(2006)Grabner, Grabner, and Bischof}]{grabner2006real}
\bibinfo{author}{H.~Grabner}, \bibinfo{author}{M.~Grabner},
  \bibinfo{author}{H.~Bischof}, \bibinfo{title}{Real-Time Tracking via On-line
  Boosting}, in: \bibinfo{booktitle}{Proceedings of the British Machine Vision
  Conference}, vol.~\bibinfo{volume}{1}, \bibinfo{publisher}{BMVA Press},
  \bibinfo{pages}{47--56}, \bibinfo{year}{2006}.

\bibitem[{Klein and Cremers(2011)}]{klein2011boosting}
\bibinfo{author}{D.~A. Klein}, \bibinfo{author}{A.~B. Cremers},
  \bibinfo{title}{Boosting scalable gradient features for adaptive real-time
  tracking}, in: \bibinfo{booktitle}{Proceedings of IEEE International
  Conference on Robotics and Automation}, \bibinfo{publisher}{IEEE},
  \bibinfo{pages}{4411--4416}, \bibinfo{year}{2011}.

\bibitem[{Zhang and Song(2013)}]{zhang2013owmil}
\bibinfo{author}{K.~Zhang}, \bibinfo{author}{H.~Song},
  \bibinfo{title}{Real-time visual tracking via online weighted multiple
  instance learning}, \bibinfo{journal}{Pattern Recognition}
  \bibinfo{volume}{46}~(\bibinfo{number}{1}) (\bibinfo{year}{2013})
  \bibinfo{pages}{397 -- 411}.

\bibitem[{Tang et~al.(2007)Tang, Brennan, Zhao, and Tao}]{tang2007cotracking}
\bibinfo{author}{F.~Tang}, \bibinfo{author}{S.~Brennan},
  \bibinfo{author}{Q.~Zhao}, \bibinfo{author}{H.~Tao},
  \bibinfo{title}{Co-tracking using semi-supervised support vector machines},
  in: \bibinfo{booktitle}{Proceedings of IEEE International Conference on
  Computer Vision}, \bibinfo{publisher}{IEEE}, \bibinfo{pages}{1--8},
  \bibinfo{year}{2007}.

\bibitem[{Hare et~al.(2011)Hare, Saffari, and Torr}]{hare2011struck}
\bibinfo{author}{S.~Hare}, \bibinfo{author}{A.~Saffari}, \bibinfo{author}{P.~H.
  Torr}, \bibinfo{title}{Struck: Structured output tracking with kernels}, in:
  \bibinfo{booktitle}{Proceedings of IEEE International Conference on Computer
  Vision}, \bibinfo{publisher}{IEEE}, \bibinfo{pages}{263--270},
  \bibinfo{year}{2011}.

\bibitem[{Bengio(2013)}]{bengio2013repres}
\bibinfo{author}{Y.~Bengio}, \bibinfo{title}{Representation Learning: A Review
  and New Perspectives}, \bibinfo{journal}{IEEE Transactions on Pattern
  Analysis and Machine Intelligence} \bibinfo{volume}{35}~(\bibinfo{number}{8})
  (\bibinfo{year}{2013}) \bibinfo{pages}{1798--1828}.

\bibitem[{Netzer et~al.(2011)Netzer, Wang, Coates, Bissacco, Wu, and
  Ng}]{netzer2011reading}
\bibinfo{author}{Y.~Netzer}, \bibinfo{author}{T.~Wang},
  \bibinfo{author}{A.~Coates}, \bibinfo{author}{A.~Bissacco},
  \bibinfo{author}{B.~Wu}, \bibinfo{author}{A.~Y. Ng}, \bibinfo{title}{Reading
  digits in natural images with unsupervised feature learning}, in:
  \bibinfo{booktitle}{NIPS Workshop on Deep Learning and Unsupervised Feature
  Learning}, \bibinfo{year}{2011}.

\bibitem[{Yu et~al.(2011)Yu, Lin, and Lafferty}]{yu2011learning}
\bibinfo{author}{K.~Yu}, \bibinfo{author}{Y.~Lin},
  \bibinfo{author}{J.~Lafferty}, \bibinfo{title}{Learning image representations
  from the pixel level via hierarchical sparse coding}, in:
  \bibinfo{booktitle}{Proceedings of IEEE Conference on Computer Vision and
  Pattern Recognition}, \bibinfo{publisher}{IEEE}, \bibinfo{pages}{1713--1720},
  \bibinfo{year}{2011}.

\bibitem[{Wang et~al.(2012{\natexlab{a}})Wang, Chen, Yang, Xu, and
  Yang}]{wang2012transferring}
\bibinfo{author}{Q.~Wang}, \bibinfo{author}{F.~Chen},
  \bibinfo{author}{J.~Yang}, \bibinfo{author}{W.~Xu}, \bibinfo{author}{M.-H.
  Yang}, \bibinfo{title}{Transferring visual prior for online object tracking},
  \bibinfo{journal}{IEEE Transactions on Image Processing}
  \bibinfo{volume}{21}~(\bibinfo{number}{7})
  (\bibinfo{year}{2012}{\natexlab{a}}) \bibinfo{pages}{3296--3305}.

\bibitem[{Liu et~al.(2013)Liu, Shen, Reid, and Hengel}]{liu2013online}
\bibinfo{author}{F.~Liu}, \bibinfo{author}{C.~Shen}, \bibinfo{author}{I.~Reid},
  \bibinfo{author}{A.~v.~d. Hengel}, \bibinfo{title}{Online Unsupervised
  Feature Learning for Visual Tracking}, \bibinfo{journal}{arXiv preprint
  arXiv:1310.1690} .

\bibitem[{Jonghoon et~al.(2013)Jonghoon, Dundar, Bates, Farabet, and
  Culurciello}]{jonghoon2013tracking}
\bibinfo{author}{J.~Jonghoon}, \bibinfo{author}{A.~Dundar},
  \bibinfo{author}{J.~Bates}, \bibinfo{author}{C.~Farabet},
  \bibinfo{author}{E.~Culurciello}, \bibinfo{title}{Tracking with deep neural
  networks}, in: \bibinfo{booktitle}{Proceedings of Conference on Information
  Sciences and Systems}, \bibinfo{pages}{1--5}, \bibinfo{year}{2013}.

\bibitem[{Wang and Yeung(2013)}]{wang2013learning}
\bibinfo{author}{N.~Wang}, \bibinfo{author}{D.-Y. Yeung},
  \bibinfo{title}{Learning a Deep Compact Image Representation for Visual
  Tracking}, in: \bibinfo{booktitle}{Proceedings of Conference on Neural
  Information Processing Systems}, \bibinfo{pages}{809--817},
  \bibinfo{year}{2013}.

\bibitem[{Mobahi et~al.(2009)Mobahi, Collobert, and Weston}]{mobahi2009deep}
\bibinfo{author}{H.~Mobahi}, \bibinfo{author}{R.~Collobert},
  \bibinfo{author}{J.~Weston}, \bibinfo{title}{Deep learning from temporal
  coherence in video}, in: \bibinfo{booktitle}{Proceedings of International
  Conference on Machine Learning}, \bibinfo{pages}{737--744},
  \bibinfo{year}{2009}.

\bibitem[{Zou et~al.(2012)Zou, Ng, Zhu, and Yu}]{zou2012deep}
\bibinfo{author}{W.~Zou}, \bibinfo{author}{A.~Ng}, \bibinfo{author}{S.~Zhu},
  \bibinfo{author}{K.~Yu}, \bibinfo{title}{Deep learning of invariant features
  via simulated fixations in video}, in: \bibinfo{booktitle}{Proceedings of
  Conference on Neural Information Processing Systems},
  \bibinfo{pages}{3212--3220}, \bibinfo{year}{2012}.

\bibitem[{Vincent et~al.(2008)Vincent, Larochelle, Bengio, and
  Manzagol}]{vincent2008extracting}
\bibinfo{author}{P.~Vincent}, \bibinfo{author}{H.~Larochelle},
  \bibinfo{author}{Y.~Bengio}, \bibinfo{author}{P.-A. Manzagol},
  \bibinfo{title}{Extracting and composing robust features with denoising
  autoencoders}, in: \bibinfo{booktitle}{Proceedings of Iinternational
  Conference on Machine Learning}, \bibinfo{pages}{1096--1103},
  \bibinfo{year}{2008}.

\bibitem[{Rifai et~al.(2011)Rifai, Vincent, Muller, Glorot, and
  Bengio}]{rifai2011contractive}
\bibinfo{author}{S.~Rifai}, \bibinfo{author}{P.~Vincent},
  \bibinfo{author}{X.~Muller}, \bibinfo{author}{X.~Glorot},
  \bibinfo{author}{Y.~Bengio}, \bibinfo{title}{Contractive auto-encoders:
  Explicit invariance during feature extraction}, in:
  \bibinfo{booktitle}{Proceedings of International Conference on Machine
  Learning}, \bibinfo{pages}{833--840}, \bibinfo{year}{2011}.

\bibitem[{Le et~al.(2011{\natexlab{a}})Le, Zou, Yeung, and Ng}]{le2011learning}
\bibinfo{author}{Q.~V. Le}, \bibinfo{author}{W.~Y. Zou}, \bibinfo{author}{S.~Y.
  Yeung}, \bibinfo{author}{A.~Y. Ng}, \bibinfo{title}{Learning hierarchical
  invariant spatio-temporal features for action recognition with independent
  subspace analysis}, in: \bibinfo{booktitle}{Proceeedings of IEEE Conference
  on Computer Vision and Pattern Recognition}, \bibinfo{organization}{IEEE},
  \bibinfo{pages}{3361--3368}, \bibinfo{year}{2011}{\natexlab{a}}.

\bibitem[{Le et~al.(2011{\natexlab{b}})Le, Karpenko, Ngiam, and Ng}]{le2011ica}
\bibinfo{author}{Q.~V. Le}, \bibinfo{author}{A.~Karpenko},
  \bibinfo{author}{J.~Ngiam}, \bibinfo{author}{A.~Y. Ng}, \bibinfo{title}{ICA
  with reconstruction cost for efficient overcomplete feature learning}, in:
  \bibinfo{booktitle}{Proceedings of Conference on Neural Information
  Processing Systems}, \bibinfo{pages}{1017--1025},
  \bibinfo{year}{2011}{\natexlab{b}}.

\bibitem[{Glorot et~al.(2011)Glorot, Bordes, and Bengio}]{glorot2011deep}
\bibinfo{author}{X.~Glorot}, \bibinfo{author}{A.~Bordes},
  \bibinfo{author}{Y.~Bengio}, \bibinfo{title}{Deep sparse rectifier networks},
  in: \bibinfo{booktitle}{Proceedings of International Conference on Artificial
  Intelligence and Statistics}, vol.~\bibinfo{volume}{15},
  \bibinfo{pages}{315--323}, \bibinfo{year}{2011}.

\bibitem[{Bengio and Bergstra(2009)}]{bengio2009slow}
\bibinfo{author}{Y.~Bengio}, \bibinfo{author}{J.~S. Bergstra},
  \bibinfo{title}{Slow, Decorrelated Features for Pretraining Complex Cell-like
  Networks}, in: \bibinfo{booktitle}{Proceedings of Conference on Neural
  Information Processing Systems}, \bibinfo{pages}{99--107},
  \bibinfo{year}{2009}.

\bibitem[{Olshausen et~al.(2007)Olshausen, Cadieu, Culpepper, and
  Warland}]{olshausen2007bilinear}
\bibinfo{author}{B.~A. Olshausen}, \bibinfo{author}{C.~Cadieu},
  \bibinfo{author}{J.~Culpepper}, \bibinfo{author}{D.~K. Warland},
  \bibinfo{title}{Bilinear models of natural images}, in:
  \bibinfo{booktitle}{Proceedings of SPIE}, vol. \bibinfo{volume}{6492},
  \bibinfo{pages}{649206--649206--10}, \bibinfo{year}{2007}.

\bibitem[{Ngiam et~al.(2011)Ngiam, Coates, Lahiri, Prochnow, Le, and
  Ng}]{ngiam2011optimization}
\bibinfo{author}{J.~Ngiam}, \bibinfo{author}{A.~Coates},
  \bibinfo{author}{A.~Lahiri}, \bibinfo{author}{B.~Prochnow},
  \bibinfo{author}{Q.~V. Le}, \bibinfo{author}{A.~Y. Ng}, \bibinfo{title}{On
  optimization methods for deep learning}, in: \bibinfo{booktitle}{Proceedings
  of International Conference on Machine Learning}, \bibinfo{pages}{265--272},
  \bibinfo{year}{2011}.

\bibitem[{Larochelle et~al.(2009)Larochelle, Bengio, Louradour, and
  Lamblin}]{larochelle2009exploring}
\bibinfo{author}{H.~Larochelle}, \bibinfo{author}{Y.~Bengio},
  \bibinfo{author}{J.~Louradour}, \bibinfo{author}{P.~Lamblin},
  \bibinfo{title}{Exploring strategies for training deep neural networks},
  \bibinfo{journal}{The Journal of Machine Learning Research}
  \bibinfo{volume}{10} (\bibinfo{year}{2009}) \bibinfo{pages}{1--40}.

\bibitem[{Coates et~al.(2011)Coates, Ng, and Lee}]{coates2011analysis}
\bibinfo{author}{A.~Coates}, \bibinfo{author}{A.~Y. Ng},
  \bibinfo{author}{H.~Lee}, \bibinfo{title}{An analysis of single-layer
  networks in unsupervised feature learning}, in:
  \bibinfo{booktitle}{Proceedings of International Conference on Artificial
  Intelligence and Statistics}, \bibinfo{pages}{215--223},
  \bibinfo{year}{2011}.

\bibitem[{Zou et~al.(2011)Zou, Ng, and Yu}]{zou2011unsupervised}
\bibinfo{author}{W.~Y. Zou}, \bibinfo{author}{A.~Y. Ng},
  \bibinfo{author}{K.~Yu}, \bibinfo{title}{Unsupervised learning of visual
  invariance with temporal coherence}, in: \bibinfo{booktitle}{NIPS Workshop on
  Deep Learning and Unsupervised Feature Learning}, \bibinfo{year}{2011}.

\bibitem[{Erhan et~al.(2009)Erhan, Bengio, Courville, and
  Vincent}]{erhan2009visualizing}
\bibinfo{author}{D.~Erhan}, \bibinfo{author}{Y.~Bengio},
  \bibinfo{author}{A.~Courville}, \bibinfo{author}{P.~Vincent},
  \bibinfo{title}{Visualizing higher-layer features of a deep network},
  \bibinfo{type}{Tech. Rep.} \bibinfo{number}{1341},
  \bibinfo{institution}{University of Montreal}, \bibinfo{year}{2009}.

\bibitem[{Wu et~al.(2013)Wu, Lim, and Yang}]{wu2013online}
\bibinfo{author}{Y.~Wu}, \bibinfo{author}{J.~Lim}, \bibinfo{author}{M.-H.
  Yang}, \bibinfo{title}{Online object tracking: A benchmark}, in:
  \bibinfo{booktitle}{Proceeedings of IEEE Conference on Computer Vision and
  Pattern Recognition}, \bibinfo{publisher}{IEEE}, \bibinfo{pages}{2411--2418},
  \bibinfo{year}{2013}.

\bibitem[{Klein et~al.(2010)Klein, Schulz, Frintrop, and
  Cremers}]{klein2010adaptive}
\bibinfo{author}{D.~A. Klein}, \bibinfo{author}{D.~Schulz},
  \bibinfo{author}{S.~Frintrop}, \bibinfo{author}{A.~B. Cremers},
  \bibinfo{title}{Adaptive real-time video-tracking for arbitrary objects}, in:
  \bibinfo{booktitle}{Proceedings of International Conference on Intelligent
  Robots and Systems}, \bibinfo{organization}{IEEE}, \bibinfo{pages}{772--777},
  \bibinfo{year}{2010}.

\bibitem[{Jia et~al.(2012)Jia, Lu, and Yang}]{jia2012visual}
\bibinfo{author}{X.~Jia}, \bibinfo{author}{H.~Lu}, \bibinfo{author}{M.-H.
  Yang}, \bibinfo{title}{Visual tracking via adaptive structural local sparse
  appearance model}, in: \bibinfo{booktitle}{Proceedings of IEEE Conference on
  Computer Vision and Pattern Recognition}, \bibinfo{organization}{IEEE},
  \bibinfo{pages}{1822--1829}, \bibinfo{year}{2012}.

\bibitem[{Jain and Nagel(1979)}]{jain1979analysis}
\bibinfo{author}{R.~Jain}, \bibinfo{author}{H.-H. Nagel}, \bibinfo{title}{On
  the analysis of accumulative difference pictures from image sequences of real
  world scenes}, \bibinfo{journal}{IEEE Transactions on Pattern Analysis and
  Machine Intelligence} \bibinfo{volume}{1}~(\bibinfo{number}{2})
  (\bibinfo{year}{1979}) \bibinfo{pages}{206--214}.

\bibitem[{Laptev(2005)}]{laptev2005on}
\bibinfo{author}{I.~Laptev}, \bibinfo{title}{On Space-Time Interest Points},
  \bibinfo{journal}{International Journal of Computer Vision}
  \bibinfo{volume}{64}~(\bibinfo{number}{2-3}) (\bibinfo{year}{2005})
  \bibinfo{pages}{107--123}, ISSN \bibinfo{issn}{0920-5691}.

\bibitem[{Lazebnik et~al.(2006)Lazebnik, Schmid, and
  Ponce}]{lazebnik2006beyond}
\bibinfo{author}{S.~Lazebnik}, \bibinfo{author}{C.~Schmid},
  \bibinfo{author}{J.~Ponce}, \bibinfo{title}{Beyond bags of features: Spatial
  pyramid matching for recognizing natural scene categories}, in:
  \bibinfo{booktitle}{Proceeedings of IEEE Conference on Computer Vision and
  Pattern Recognition}, \bibinfo{publisher}{IEEE}, \bibinfo{pages}{2169--2178},
  \bibinfo{year}{2006}.

\bibitem[{Fan et~al.(2010)Fan, Xu, Wu, and Gong}]{fan2010human}
\bibinfo{author}{J.~Fan}, \bibinfo{author}{W.~Xu}, \bibinfo{author}{Y.~Wu},
  \bibinfo{author}{Y.~Gong}, \bibinfo{title}{Human tracking using convolutional
  neural networks}, \bibinfo{journal}{IEEE Transactions on Neural Networks}
  \bibinfo{volume}{21}~(\bibinfo{number}{10}) (\bibinfo{year}{2010})
  \bibinfo{pages}{1610--1623}.

\bibitem[{Gordon et~al.(1993)Gordon, Salmond, and Smith}]{gordon1993novel}
\bibinfo{author}{N.~Gordon}, \bibinfo{author}{D.~Salmond},
  \bibinfo{author}{A.~F.~M. Smith}, \bibinfo{title}{Novel approach to
  nonlinear/non-Gaussian Bayesian state estimation}, \bibinfo{journal}{IEE
  Proceedings F Radar and Signal Processing}
  \bibinfo{volume}{140}~(\bibinfo{number}{2}) (\bibinfo{year}{1993})
  \bibinfo{pages}{107--113}.

\bibitem[{Arulampalam et~al.(2002)Arulampalam, Maskell, Gordon, and
  Clapp}]{arulampalam2002a}
\bibinfo{author}{M.~Arulampalam}, \bibinfo{author}{S.~Maskell},
  \bibinfo{author}{N.~Gordon}, \bibinfo{author}{T.~Clapp}, \bibinfo{title}{A
  tutorial on particle filters for online nonlinear/non-Gaussian Bayesian
  tracking}, \bibinfo{journal}{IEEE Transactions on Signal Processing}
  \bibinfo{volume}{50}~(\bibinfo{number}{2}) (\bibinfo{year}{2002})
  \bibinfo{pages}{174--188}.

\bibitem[{Wang et~al.(2011)Wang, Lu, Yang, and Yang}]{shu2011superpixel}
\bibinfo{author}{S.~Wang}, \bibinfo{author}{H.~Lu}, \bibinfo{author}{F.~Yang},
  \bibinfo{author}{M.-H. Yang}, \bibinfo{title}{Superpixel tracking}, in:
  \bibinfo{booktitle}{Proceedings of IEEE International Conference on Computer
  Vision}, ISSN \bibinfo{issn}{1550-5499}, \bibinfo{pages}{1323--1330},
  \bibinfo{year}{2011}.

\bibitem[{Babenko et~al.(2009)Babenko, Yang, and Belongie}]{babenko2009visual}
\bibinfo{author}{B.~Babenko}, \bibinfo{author}{M.-H. Yang},
  \bibinfo{author}{S.~Belongie}, \bibinfo{title}{Visual tracking with online
  multiple instance learning}, in: \bibinfo{booktitle}{Proceedings of IEEE
  Conference on Computer Vision and Pattern Recognition},
  \bibinfo{publisher}{IEEE}, \bibinfo{pages}{983--990}, \bibinfo{year}{2009}.

\bibitem[{Zhang et~al.(2012)Zhang, Zhang, and Yang}]{zhang2012real}
\bibinfo{author}{K.~Zhang}, \bibinfo{author}{L.~Zhang}, \bibinfo{author}{M.-H.
  Yang}, \bibinfo{title}{Real-time compressive tracking}, in:
  \bibinfo{booktitle}{Proceedings of European Conference on Computer Vision},
  \bibinfo{publisher}{Springer}, \bibinfo{pages}{864--877},
  \bibinfo{year}{2012}.

\bibitem[{Zhang et~al.(2013)Zhang, Zhang, and Yang}]{zhang2013realtime}
\bibinfo{author}{K.~Zhang}, \bibinfo{author}{L.~Zhang}, \bibinfo{author}{M.-H.
  Yang}, \bibinfo{title}{Real-Time Object Tracking Via Online Discriminative
  Feature Selection}, \bibinfo{journal}{IEEE Transactions on Image Processing}
  \bibinfo{volume}{22}~(\bibinfo{number}{12}) (\bibinfo{year}{2013})
  \bibinfo{pages}{4664--4677}.

\bibitem[{Wang et~al.(2012{\natexlab{b}})Wang, Chen, Xu, and
  Yang}]{wang2012object}
\bibinfo{author}{Q.~Wang}, \bibinfo{author}{F.~Chen}, \bibinfo{author}{W.~Xu},
  \bibinfo{author}{M.-H. Yang}, \bibinfo{title}{Object tracking via partial
  least squares analysis}, \bibinfo{journal}{IEEE Transactions on Image
  Processing} \bibinfo{volume}{21}~(\bibinfo{number}{10})
  (\bibinfo{year}{2012}{\natexlab{b}}) \bibinfo{pages}{4454--4465}.

\bibitem[{Wang et~al.(2013)Wang, Lu, and Yang}]{wang2013online}
\bibinfo{author}{D.~Wang}, \bibinfo{author}{H.~Lu}, \bibinfo{author}{M.-H.
  Yang}, \bibinfo{title}{Online object tracking with sparse prototypes},
  \bibinfo{journal}{IEEE Transactions on Image Processing}
  \bibinfo{volume}{22}~(\bibinfo{number}{1}) (\bibinfo{year}{2013})
  \bibinfo{pages}{314--325}.

\bibitem[{Kalal et~al.(2012)Kalal, Mikolajczyk, and Matas}]{zdenek2012tracking}
\bibinfo{author}{Z.~Kalal}, \bibinfo{author}{K.~Mikolajczyk},
  \bibinfo{author}{J.~Matas}, \bibinfo{title}{Tracking-Learning-Detection},
  \bibinfo{journal}{IEEE Transactions on Pattern Analysis and Machine
  Intelligence} \bibinfo{volume}{34}~(\bibinfo{number}{7})
  (\bibinfo{year}{2012}) \bibinfo{pages}{1409--1422}, ISSN
  \bibinfo{issn}{0162-8828}.

\bibitem[{Vedaldi and Fulkerson(2010)}]{vedaldi2010vlfeat}
\bibinfo{author}{A.~Vedaldi}, \bibinfo{author}{B.~Fulkerson},
  \bibinfo{title}{VLFeat: An open and portable library of computer vision
  algorithms}, in: \bibinfo{booktitle}{Proceedings of the International
  Conference on Multimedia}, \bibinfo{organization}{ACM},
  \bibinfo{pages}{1469--1472}, \bibinfo{year}{2010}.

\end{thebibliography}

\footnotesize

\end{document}